\documentclass[review]{elsarticle}

\pdfoutput=1

\journal{Computer Vision and Image Understanding}

\usepackage{graphicx}
\usepackage[usenames,dvipsnames]{color}
\usepackage{amsmath,amssymb} 
\usepackage{lineno,hyperref}
\modulolinenumbers[5]
\usepackage{color}
\usepackage{times}
\usepackage{epsfig}
\usepackage{subfigure} 
\usepackage{algorithm,algorithmic}
\usepackage{multirow}
\usepackage[english]{babel}
\usepackage{tabularx}
\usepackage{multirow}
\usepackage{xspace}
\usepackage{booktabs}
\usepackage{picins}
\usepackage{epstopdf}

\newcommand{\comment}[1]{}

\newcommand{\re}{\text{re}}
\newcommand{\sy}{\text{sy}}
\newcommand{\Eucl}{\text{Eucl}}
\newcommand{\HOG}{\text{HoG}}
\newcommand{\WL}{\text{WL}}
\newcommand{\CNN}{\text{CNN}}

\DeclareMathOperator*{\argmin}{\text{argmin}}

\newcommand{\Intrinsic}{Capture\xspace}
\newcommand{\intrinsic}{capture\xspace}
\newcommand{\extrinsic}{pose\xspace}

\newcommand{\vincent}[1]{#1}
\newcommand{\vincentrmk}[1]{}
\newcommand{\artem}[1]{#1}
\newcommand{\pascal}[1]{#1}

\newcommand{\threeim}{1.4in}

\begin{document}

\begin{frontmatter}

\title{On Rendering Synthetic Images for Training an Object Detector}

\author[mymainaddress]{Artem Rozantsev\corref{mycorrespondingauthor}}
\cortext[mycorrespondingauthor]{Corresponding author}
\ead{artem.rozantsev@epfl.ch}
\ead[url]{http://cvlab.epfl.ch/~rozantse, Phone: +41(0)78 9472721}
\author[mymainaddress,mysecondaryaddress]{Vincent Lepetit}
\ead{lepetit@icg.tugraz.at}
\ead[url]{http://www.icg.tugraz.at/Members/lepetit}
\author[mymainaddress]{Pascal Fua}
\ead{pascal.fua@epfl.ch}
\ead[url]{http://cvlab.epfl.ch/~fua}

\address[mymainaddress]{\'{E}cole Polytechnique F\'{e}d\'{e}rale de Lausanne, Computer Vision Laboratory, Lausanne, Switzerland}
\address[mysecondaryaddress]{ Graz University of Technology, Institute for Computer Graphics and Vision, Graz, Austria}




\begin{abstract}

We  propose a  novel  approach to  synthesizing images  that  are effective  for
training  object detectors.   Starting  from a  small set  of  real images,  our
algorithm  estimates the  rendering  parameters required  to synthesize  similar
images given a coarse  3D model of the target object.  These parameters can then
be reused to  generate an unlimited line  of training images of  the object of
interest  in   arbitrary  3D  poses,  which   can  then  be  used   to  increase
classification performances.

A key insight of our approach  is that the synthetically generated images should
be similar to real images, not in terms of image quality, but rather in terms of
features used during the detector training.  We show in the context of
  drone, plane, and car detection that using such synthetically generated images
  yields significantly better performances than simply perturbing real images or
  even synthesizing  images in  such way  that they look  very realistic,  as is
  often done when only limited amounts of training data are available.

\end{abstract}


\end{frontmatter}



\section{Introduction}

It is now widely accepted that when enough  training data is  available,  statistical  approaches  can  address  image  classification
problems~\cite{Le12} very effectively.   In the commercial world, this  is a key
ingredient of high performing face detection software deployed by companies such
as  Apple and  Google.  However,  there are  real-world scenarios  in  which the
required training data  is hard to obtain in  sufficiently large quantities. For
example,  our  work  is motivated  by  the  emerging  need for  Unmanned  Aerial
Vehicles~(UAVs), or  \emph{drones}, to see and  avoid each other  as they become
increasingly numerous and autonomous in  the sky.  In this application, training
videos are rare and  do not cover the full range of  possible shapes, poses, and
lighting conditions under which they can be seen.

\begin{figure}
  \centering
  \includegraphics[width=2.7in]{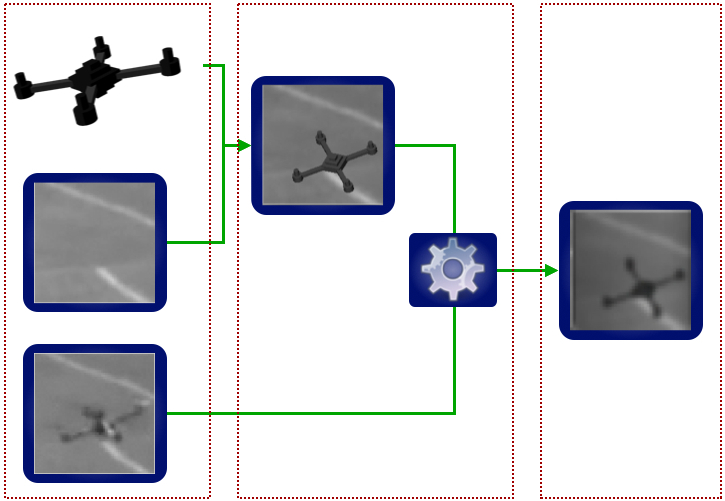} \\
  (a) $\qquad \qquad \verb"    " $ (b) $\qquad \qquad \verb"    " $ (c)
  \caption{Synthetic data generation pipeline. (a)  Input to the system includes
    a simple  model of the  object of  interest, an image  of this object  and a
    background  image (it  should be  the same  background as  the image  of the
    object has). (b)  Overlaying the model on the background  yields a synthetic
    image. This  image is then  processed to maximize  its similarity to  a real
    image  from the  perspective of  the detector.  (c) The  resulting synthetic
    image can be used for training the detector.}
  \label{fig:im_synth}
\end{figure}

Our goal  therefore is to supplement  a small number of  available real training
samples  with an  arbitrary  large  dataset of  synthetic  ones  to improve  the
detection  accuracy  of a  final  classifier.   Using  synthetic data  has  been
spectacularly   successful  for   3D   body  pose   estimation   with  a   depth
camera~\cite{Shotton12}.  However, depth data do  not vary with lighting, motion
blur, and  other artifacts  that affect  images from a  regular camera,  and are
therefore comparatively simpler to synthesize.

A training set  can also be augmented by applying  small deformations and adding
noise to  the images it  contains~\cite{Burges97,Decoste02}.  This was  done for
character~\cite{LeCun98,Varga03},   face~\cite{Fleuret01},   and   image   patch
recognition~\cite{Lepetit05a}.  Such  augmentations are typically  necessary for
the now  popular Convolutional Neural  Networks~\cite{Serre07}, which require 
large  amounts of training  data.  

However, this approach assumes that the original training set is already diverse
enough,  as the  range of  synthetic  images that  can be  produced is  limited.
Moreover simple  perturbations are often  not enough  and special car  should be
taken.   More  sophisticated  approaches  have  also  been  proposed  for  human
detection  and  pose  estimation  purposes  in~\cite{Marin10,Pishchulin12},  but
\cite{Marin10}      does     not      model     image-acquisition      artifacts
while~\cite{Pishchulin12} involves  considerable amounts of  manual interaction,
which  is less  desirable.  It  was recently  shown~\cite{Rematas14} that  it is
possible to use a 3D car model to first extract appearance information from real
images of cars, and use this information to synthesize novel views.  Using these
images for  training purposes  improves performance but  this approach  does not
account  for other  artifacts such  as  motion blur  and is  only applicable  to
objects with relatively simple geometry.

Furthermore, to  the best  of our  knowledge none of  these approaches  offers a
principled way to choose the image synthesis parameters to match the behavior of
real-world  cameras in  the  presence  of noise.   The  relevant parameters  are
typically tuned by  hand, which quickly becomes unmanageable  when the rendering
pipeline is complex. To overcome this limitation, we therefore introduce a fully
automated and  generic method to estimate  these parameters from a  small set of
available real images to maximize  the performance of a detector trained using
the resulting synthetic images.

To this end, we start from a small set of real seed images containing a target
object and corresponding background images without it, such as the ones depicted
by Fig.~\ref{fig:im_synth}(a).  Given a  very coarse 3D  model of the  object of
interest, such as  that of the drone of  Fig.~\ref{fig:im_synth}(a), we estimate
the  3D  pose of  the  object,  overlaid onto  the  background  image, and  then
post-process the resulting composite image so  that it is as similar as possible
to the real one.  This is achieved by automated selection of the post-processing
parameters  to  maximize  a  similarity  between  the  two  images.  Once  these
parameters are found, we can then change the position and the orientation of the
object  in  the  images  to  generate arbitrary  large  synthetic  datasets  with
realistic imaging artifacts.

A key ingredient of our approach is  the similarity function used to measure the
difference between real and composite images.  An obvious candidate would be the
pixel-wise Euclidean distance.   However, our goal is  not to generate eye-pleasant
images,  but rather  training data that  is effective  for our intended purpose. 
We will therefore show that the best similarity depends on the target detection
method. We demonstrate this for three widely used methods that are representative of the
state-of-the-art: The Deformable  Part Model~(DPM) method~\cite{Felzenszwalb08},
an   AdaBoost-based   detector~\cite{Freund95},   and  a   detector   based   on
Convolutional  Neural  Networks~(CNN)~\cite{Serre07}. \vincent{Together, these  methods  cover
the state-of-the-art in both object detection and image features.}

In short, our contribution is a novel and fully automated approach to generating
synthetic  training image  databases  that increases  detection performance  and
outperforms the state-of-the-art techniques discussed above, irrespective of the
specific detector  used.  We will  demonstrate this  in the context  of drone,
plane, and car detection.

In the remainder of this paper, we first discuss the effects we want to model in
our synthetic  images.  We  then describe and  compare the  different similarity
functions  to  quantify  the  similarity  between  synthetic  and  real  images.
Furthermore,  we demonstrate  the power  of our  approach on  aircrafts of  very
different   shapes   and   flying   in   various   environments   and   lighting
conditions. Finally,  we compare our  approach to  recent work~\cite{Rematas14}
on the Pascal VOC dataset.



\section{Related Work}
\label{sec:related}

Given  the  prevalence  of  Machine  Learning based  algorithms,  capturing  and
annotating  training images has  become a  major issue,  and sometimes  a severe
bottleneck when  such images are hard  to acquire. In such  cases, using Computer
Graphics techniques to generate them is a very attractive alternative.

For example, Optical Character Recognition  systems have long been trained using
samples  created by  applying various  deformations  and adding  image noise  to
actual samples~\cite{LeCun98,Varga03}.  Similarly, synthetically generated image
patches  have  been  successfully  used  in~\cite{Lepetit06a,Ciresan12b}.  Note,
however,  that neither  characters nor  patches exhibit  the full  complexity of
natural images and are therefore easier to synthesize. In~\cite{Shotton12}, this
approach was used on complete depth images generated from 3D models of people to
train classifiers to recover  human 3D pose from the output  of a Kinect camera.
This has been remarkably successful, in large  part because it provides a way to
create arbitrarily large training dataset.  However, depth images also lack many
of the  imaging artefacts  present in  ordinary images, such  as motion  blur or
lighting effects,  which make  it difficult  to use such  an approach  for video
imagery.

This  was attempted  in~\cite{Marin10} by  generating images  of pedestrians  in
various poses and environments to train  a pedestrian detector.  The results are
encouraging but the method does not take complex imaging artefacts into account.
More  recently, an  approach  to  creating more  realistic  synthetic images  by
extracting people's  silhouettes from real  images, and superimposing  them over
various  backgrounds  was  proposed~\cite{Pishchulin12}.  However,  it  is  very
specific to  pedestrian detection and  requires a considerable amount  of manual
annotation.

Like ours, the approach of~\cite{Rematas14}  relies on both real training images
and a 3D model.  After registering the  3D model to the images, the material and
lighting properties of the different object components are estimated and used to
synthesise new  views of the  3D model. However, it  does not take  into account
other artifacts such as motion blur and requires precise registration.

Of course,  generic image synthesis techniques  have also been used  in computer
vision   for  many   other  purposes,   such  as   optimizing  camera   tracking
algorithms~\cite{Handa12},                     evaluation                     of
algorithms~\cite{Horn81,Barron94,Stark10,Liebelt10,Baker11,Kaneva11},    gesture
recognition  and   pose  estimation~\cite{Athitsos03,Taycher06},   or  rendering
virtual objects that merge well  with real images~\cite{Klein10}.  Some of these
approaches simply project the 3D model of the object of interest on an arbitrary
background image.   Others add post-processing on  similarly generated synthetic
images  in order  to make  them look  realistic.  However,  to the  best of  our
knowledge none of them estimate neither  how realistic the resulting images are,
nor how suitable they are for the application itself.

In this work, we will use some  of the same approaches to synthesizing realistic
images. This  being said,  visual realism is  not our end  goal, but  rather the
classification  performance  improvement.  As  such our  algorithm,  unlike  the
others,  automatically  optimizes  the  rendering  parameters  solely  for  this
purpose.



\section{Generating Synthetic Images}
\label{sec:synth_data}

As  illustrated by  Fig.~\ref{fig:im_synth}, while  our pipeline  is simple,  it
depends on many parameters  that would be hard to choose by  hand.  We use simple
CAD  models, such  as that  of Fig.~\ref{fig:im_synth}(a),  which roughly
captures the target object geometry.  We assume that we are given a small set of
real images  featuring the target object  and a corresponding set  of background
images without  it.  \vincent{As  we will explain,  these background  images can
  usually be extracted  from the training video sequence itself.  In cases where
  the background is not visible at any time, it is still possible to estimate it
  by cutting out the object from the original images and using a texture filling
  algorithm.   This   approach   will    be   more   thoroughly   discussed   in
  Section~\ref{PVOC_comp}.}

For  each real image, we then compute 5 {\it  \extrinsic} parameters, that include 3 orientations $(\alpha^p, \beta^p, \gamma^p)$ and 2 translations $(t^p_x, t^p_y)$, which lets us project the 3D model at the desired location. Note that as we use multi-scale detector, we do not need to vary the scale of the object.

\begin{figure}
\centering
\begin{tabular}{ccc}
\includegraphics[width=2.0in]{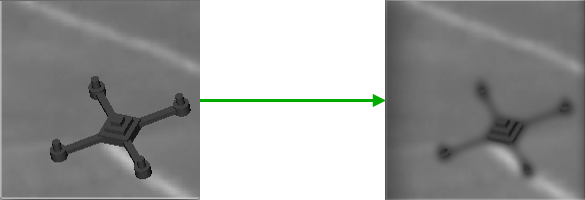}  & & \includegraphics[width=2.0in]{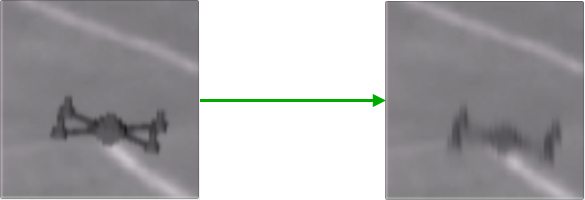}\\
(a) Boundaries blurring & & (b) Motion blurring \\
\includegraphics[width=2.0in]{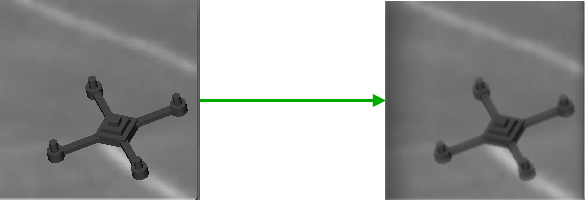} & & \includegraphics[width=2.0in]{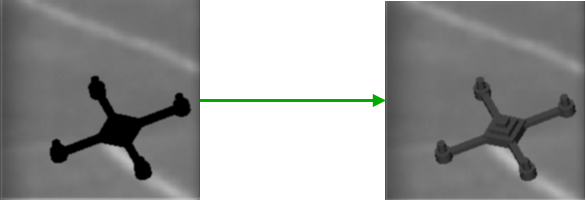}\\
(c) Random noise & & (d) Diffuse coefficient of material variation \\
\end{tabular}
\caption{The four post-processing  effects we use for  increasing the similarity
  between the real and synthetic images.}
\label{fig:effects}
\end{figure}

As shown in Fig.~\ref{fig:effects}, we  then post-process the synthetic image to
maximize its similarity to the real image. This involves:
\begin{itemize}

\item \emph{Object  boundary blurring} {\bf  (BB)}.  The discrete nature  of the
  image sensor causes a mixture  of the intensities of  the background
  and the target object along its  boundaries.  To simulate this effect we apply
  Gaussian blurring along the object boundaries  after the object image has been
  overlaid  on  the  background  image.   This is  controlled  by  the  standard
  deviation $\sigma^s$ of the Gaussian kernel used for smoothing.

\item \emph{Motion  blurring} {\bf (MB)}.  This mimics the blurring  effect that
  affects  on fast  moving objects  if the  shutter time  of the  camera is  too
  long. To simulate this effect we  use anisotropic Gaussian blurring applied to
  the pixels of the  object in the direction of its  motion.  The parameters are
  the  two standard  deviations $\sigma^m_u$  and $\sigma^m_v$  of the  Gaussian
  kernel and the angle $\alpha^m$ of the motion.

\item \emph{Random noise} {\bf (RN)}.   This emulates the shot noise added to
  the image  by the camera.  To  simulate this effect we  simply add independent
  Gaussian noise  to the pixel  intensities. Note this  is limited to  the image pixels
  that correspond to the inserted object, as the  background images are real
  ones and  already contain similar noise.  This is controlled by the standard
  deviation $\sigma^n$ of the Gaussian distribution used to generate the noise.
\item  \emph{Material properties} {\bf (MP)}.   We also  vary the  material properties,  by
  changing the  weight $w^d$ of the  diffuse reflection.  This  allows us not  only to
  vary the color of the object, but also to introduce some diffuse lighting effects. While we do not take specularities into account, this would be a very natural extension to our approach.
\end{itemize}
We refer to these synthetic data generation parameters as {\it
  \intrinsic} parameters
\begin{equation}
\Theta  = [\underbrace{\alpha^p, \beta^p, \gamma^p,  t^p_x, t^p_y}_\text{\extrinsic}, \underbrace{\sigma^s,
  \sigma^m_u,  \sigma^m_v, \alpha^m,  \sigma^n, w^d}_\text{\intrinsic}]^\top \> .
\label{eq:Intrinsic}
\end{equation}
These parameters  are challenging to  tune because they are  heavily correlated.
This is particularly  true of object pose and direction of  motion blur, as well
of boundary blurring and motion blurring.
Thus our goal  is to estimate the $\Theta$ parameters for  every seed real image
that we use for synthetic data  generation.  Given the background images and the
corresponding  $\Theta$  parameters, we  retain  the  \intrinsic parameters  and
randomize the \extrinsic  ones to generate arbitrary large  numbers of synthetic
images  that  will be  realistic  enough  to be  used  for  training the  object
detector.  We explain below how we recover these $\Theta$ parameters.


\section{Optimizing the Rendering Parameters}
\label{sec:optimizing}

To optimize the \extrinsic  and   \intrinsic parameters in $\Theta$,  we rely on a small set  of real images of
the target  object, together with the corresponding  images of the background  without the
target.

Starting from a background image on which  we render the CAD model of the target
object, we optimize the rendering parameters to reproduce the corresponding real
image. This optimization is performed on each image independently,
because the same capture parameters do not necessarily apply to all of
them.  More formally, we consider the set of pairs of real images $\{  (  X_i, B_i)  \}^N_{i=0}$, where $X_i \in$ {\Large$\chi$} is the $i^\text{th}$ image of the object and $B_i \in$ {\Large$\chi$} is the background image for $X_i$. Let $d: ${\Large$\chi$} $\times$ {\Large$\chi$} $\rightarrow \mathbf{R}^+$ be a similarity function, which we use to compare two images, and which we will define explicitly in Sect.~\ref{subsec:sim_measure}. Lastly, let $S(\Theta,B_i) \in $ {\Large$\chi$} represent the synthetically rendered image by applying the synthetic data generation process with parameters $\Theta$ to the Background $B_i$

To find the set of parameters $\Theta$ that best corresponds to
real image $X_i$, we look for
\begin{equation}
\Theta^{(i)} = \argmin_\Theta d(X_i, S(\Theta, B_i)) 
\label{eq:problem_math}
\end{equation}
by Simulated  Annealing~\cite{Kirkpatrick83}. \vincent{This approach is
  widely used for solving non-continuous optimization problems with a
  large number of parameters.} In practice, we initialize the \extrinsic parameters by manually providing the object center, which could be avoided with a more sophisticated optimization algorithm. \Intrinsic parameters are initialized randomly. This optimization takes a few seconds on each of our $40 \times 40$ images.

\begin{figure*}
\centering
\begin{tabular}{m{0.06in}m{1.2in}m{0.06in}m{1.2in}m{0.06in}m{1.2in}}
 $\sigma^n$ &
\includegraphics[width=1.4in]{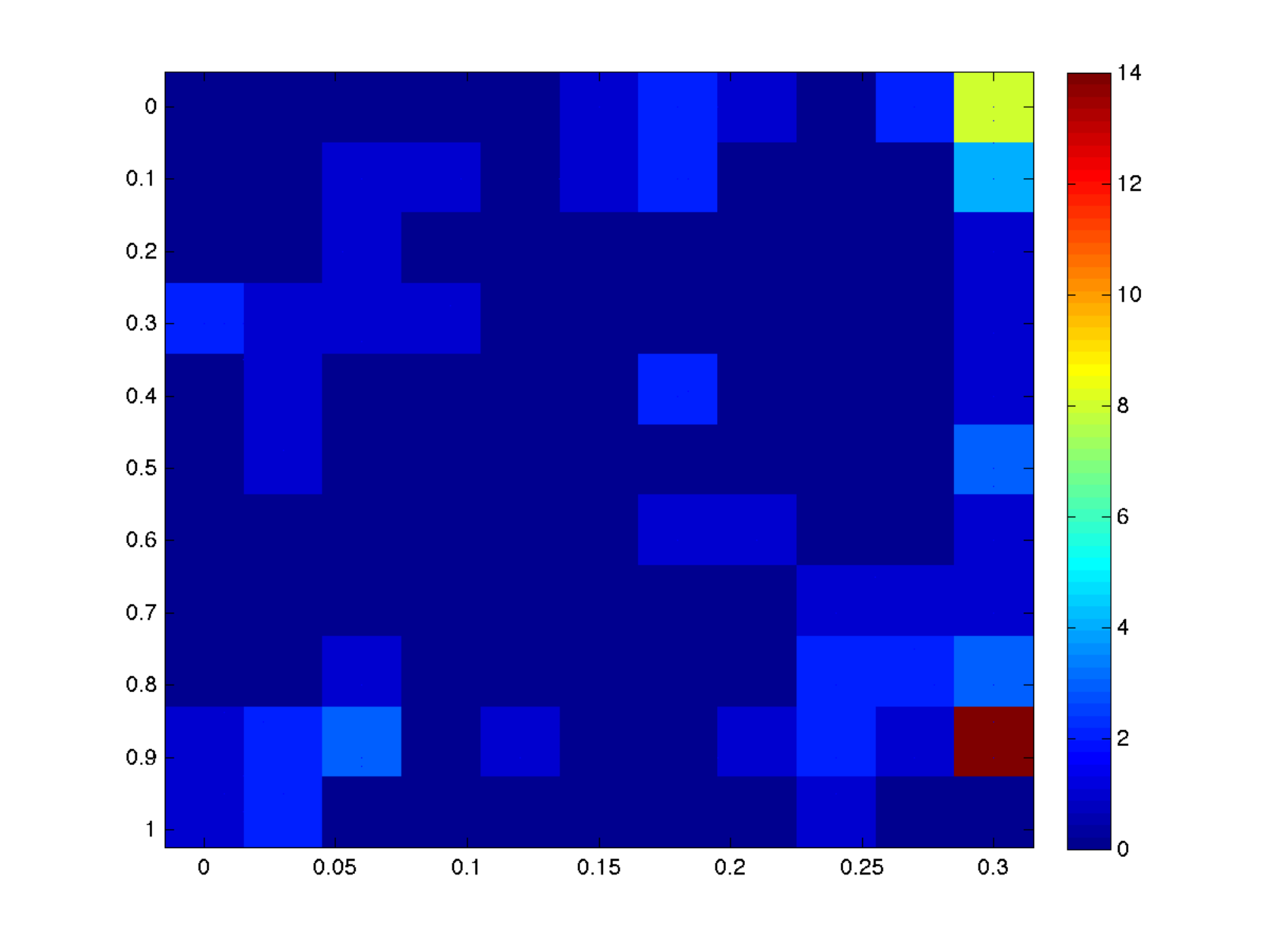} &
 $\sigma^n$ &
 \includegraphics[width=1.4in]{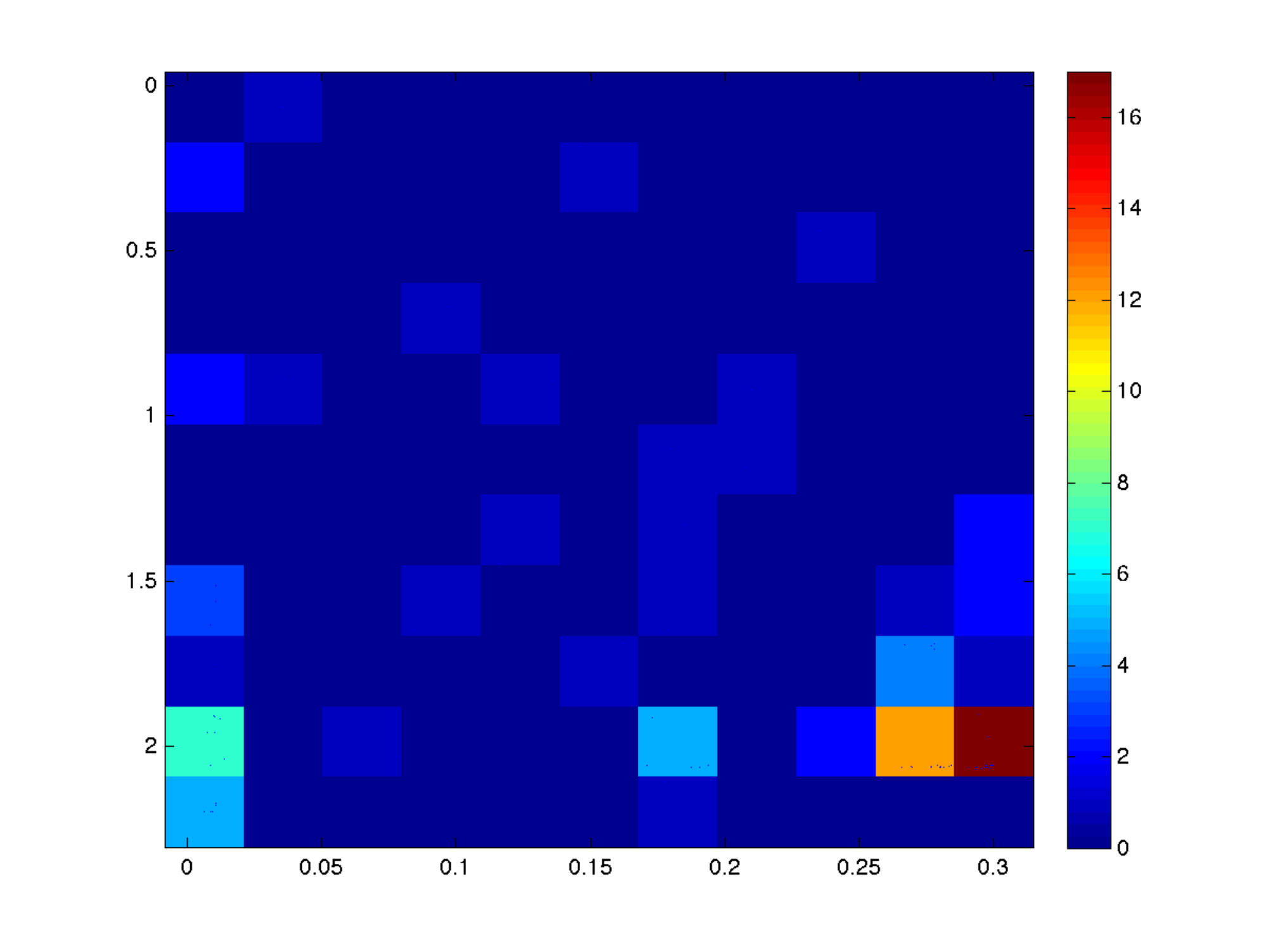} &
  $\sigma^m_u$ &
 \includegraphics[width=1.4in]{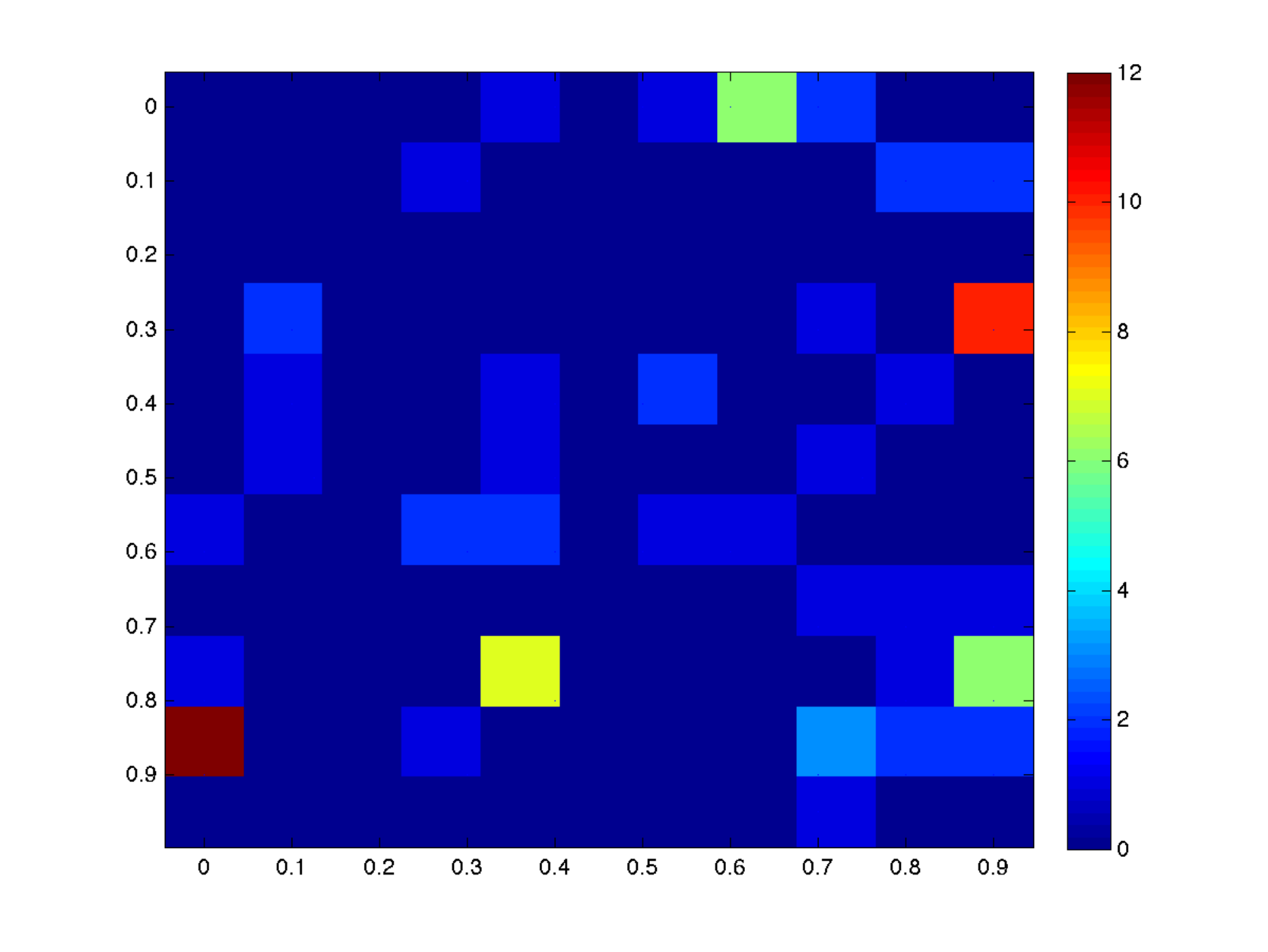} \\
 & \multicolumn{1}{c}{$\sigma^m_v$} & & \multicolumn{1}{c}{$\sigma^s$} & & \multicolumn{1}{c}{$\sigma^m_v$} \\
\end{tabular}
\caption{Each histogram  depicts the joint distribution  of different pairs of \intrinsic
  parameters. (best seen in color)}
  \label{fig:Distr_par}
\end{figure*}

 
The  \intrinsic parameters  in $\Theta$  depend on  viewing conditions,  such as
lighting and  weather conditions, which  is why  we perform the  optimization in
each    image   independently.     Fig.~\ref{fig:Distr_par}   describes    their
distributions across images.   Note that these distributions  are absolutely not
Gaussian  and  that   it  would  therefore  be  non-trivial   to  describe  them
analytically.  \comment{Full set  of distributions between different  pairs of parameters is presented in the supplementary material.}

\subsection{Image Similarity Measures}
\label{subsec:sim_measure}

The  resulting   parameters  depend   critically  on  the   similarity  function
$d(\cdot,\cdot)$ used  to evaluate how close  the two images are  to each other.
The simplest  is the  Euclidean distance between the  intensity values of
corresponding pixels
\begin{equation}\label{eq:euclid_dist}
d_\Eucl(X_\re,X_\sy)     =    \sqrt{    \mathop{\Sigma}\limits_{v     =    1}^H
  \mathop{\Sigma}\limits_{u = 1}^W \big(X_\re(u,v) - X_\sy(u,v)\big)^2 } \> ,
\end{equation}
where  $X_\re$  and   $X_\sy$  are  the  real   and  synthetic  images
respectively, and $W$ and $H$ denote the images dimensions.

However, since our goal  is to generate synthetic images that  are more
effective to train a detection method, we will see this is not the best possible choice, for our purposes.

More specifically, we evaluated our approach in conjunction  with three commonly
used    object    detectors---DPM~\cite{Felzenszwalb08},    an    AdaBoost-based
detector~\cite{Freund95}, and a  CNN~\cite{Serre07}---and we therefore introduce
three different similarity functions, each one based on the image features used by
one  of these  methods. We will  show that  our approach  to image
generation works best when relying on  the distance function  corresponding to
the detection method.

Since  DPM  relies  on Histograms-of-Gradients~(HoG)~\cite{Dalal05},  the  first
similarity function  we  consider  the  distance  between  the  HoG  vectors~\cite{Dalal05}
computed for the two images as the similarity function
\begin{equation}\label{eq:HoG_dist}
d_\HOG(X_\re,X_\sy)  =  \sqrt{ \mathop{\Sigma}  \limits_{i  =  1}^L
\big(\HOG_i(X_\re) - \HOG_i(X_\sy)\big)^2} \> ,
\end{equation}
where  $\HOG_i(X)$ is the $i^\text{th}$ coordinate of the HoG vector computed
for image $X$.

We also consider an AdaBoost detector, whose weak  learners rely on the  image gradients
proposed in~\cite{Levi04}. We write
\begin{equation}
\label{eq:wl}
h_{R,o,\tau}(X) = \left \{
   \begin{array}{ll}
       1, & \text{ if } E(X, R, o) > \tau, \\ 
       0, & \text{ otherwise.} \\
   \end{array}
   \right. 
\end{equation}
These weak learners are parametrized by a region $R$, an orientation $o$, and a
threshold $\tau$.   $E(X, R, e)$  is the  normalized image gradient  energy over
region $R$  in $X$  and in  orientation $o$.  We therefore introduce the additional function:
\begin{equation}\label{eq:wl_dist}
d_\WL^H(X_\re,X_\sy)  =  \sqrt{ \mathop{\Sigma}  \limits_{i = 1} \limits^{L} \alpha_{i} \big(h_i(X_\re) - h_i(X_\sy)\big)^2} \> ,
\end{equation}
where $L$ is the number of  weak learners $h_i$ with their corresponding weights
$\alpha_i$. We tried two different methods to build such a set:
\begin{itemize}
\item $d_\WL^R(X_\re,X_\sy)$ will denote the previous similarity function when random weak
learners, each with a weight $\alpha = 1$, are used;
\item $d_\WL^L(X_\re,X_\sy)$ will denote the previous similarity function when a set of weak
learners and their weights selected by AdaBoost on the seed real images is used.
\end{itemize}

The third detection  method we consider is a  Convolutional Neural Network~(CNN)
which, unlike the previous  two, does not rely  on hard-coded image
features but learns them instead. We therefore first train a CNN on the real
seed images only and consider the  distance
\begin{equation}\label{eq:CNN_dist}
d_\CNN(X_\re,X_\sy)  =  \sqrt{ \mathop{\Sigma}  \limits_{n = 2}^N  \mathop{\Sigma}  \limits_{i  =  1}^{L_n}
\big(\CNN^n_i(X_\re) - \CNN^n_i(X_\sy)\big)^2} \> ,
\end{equation}
where $\CNN^n_i(X)$  is the value of the $i^\text{th}$ neuron of the
$n^\text{th}$ layer of the Convolutional Neural Network; $N$ is the number of
layers in the CNN; $L_n$ is the number of neurons of the $n^\text{th}$ layer of CNN.

\begin{figure}
\centering
\begin{tabular}{ccccccc}
Original & \hspace{2mm} & $d_\Eucl$ & $d_\HOG$ & $d_\WL^R$ & $d_\WL^L$ & $d_\CNN$  \\
\\
\includegraphics[width=0.6in]{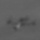} & \hfil & 
\includegraphics[width=0.6in]{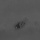} &
\includegraphics[width=0.6in]{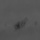} & 
\includegraphics[width=0.6in]{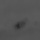} & 
\includegraphics[width=0.6in]{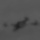} & 
\includegraphics[width=0.6in]{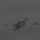} \\
\includegraphics[width=0.6in]{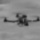} & \hfil & 
\includegraphics[width=0.6in]{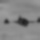} &
\includegraphics[width=0.6in]{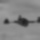} & 
\includegraphics[width=0.6in]{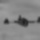} & 
\includegraphics[width=0.6in]{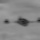} & 
\includegraphics[width=0.6in]{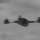} \\
\includegraphics[width=0.6in]{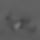} & \hfil & 
\includegraphics[width=0.6in]{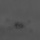} &
\includegraphics[width=0.6in]{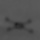} & 
\includegraphics[width=0.6in]{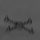} & 
\includegraphics[width=0.6in]{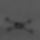} & 
\includegraphics[width=0.6in]{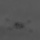} \\
\includegraphics[width=0.6in]{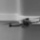} & \hfil & 
\includegraphics[width=0.6in]{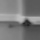} &
\includegraphics[width=0.6in]{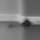} & 
\includegraphics[width=0.6in]{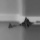} & 
\includegraphics[width=0.6in]{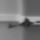} & 
\includegraphics[width=0.6in]{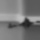} \\
\end{tabular}
\caption{Samples of real images with  corresponding synthetic ones. The $\Theta$
  parameters for the synthetic images  were optimised using different image similarity
  functions.}
\label{fig:RS_samples}
\end{figure}

In Fig.~\ref{fig:RS_samples},  we show  synthetic images with  the corresponding
real seed  images. Each image was  obtained by finding the  rendering parameters
that minimize one of the five similarity functions introduced above.  

\newcolumntype{M}[1]{>{\centering\arraybackslash}m{#1}}
\newcommand{\ra}[1]{\renewcommand{\arraystretch}{#1}}

\newcommand{\dbimsz}{0.45in}

\begin{figure}
  \centering
  \begin{tabular}{|M{0.2in}c|}
    \hline
    \rotatebox{90}{Training Data} &
    \begin{tabular}{ccccc}
      & & & & \\
      \includegraphics[width=\dbimsz]{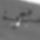} &
      \includegraphics[width=\dbimsz]{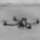} &
      \includegraphics[width=\dbimsz]{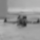} &
      \includegraphics[width=\dbimsz]{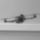} &
      \includegraphics[width=\dbimsz]{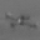} \\
      \includegraphics[width=\dbimsz]{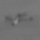} &
      \includegraphics[width=\dbimsz]{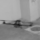} &
      \includegraphics[width=\dbimsz]{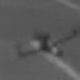} &
      \includegraphics[width=\dbimsz]{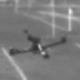} &
      \includegraphics[width=\dbimsz]{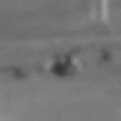} \\
      & & & & \\
    \end{tabular} \\
    \hline
    \rotatebox{90}{Evaluation Data} &
    \begin{tabular}{ccccccc}
      & & & & & & \\
      \includegraphics[width=\dbimsz]{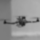}&
      \includegraphics[width=\dbimsz]{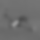} &
      \includegraphics[width=\dbimsz]{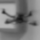} &
      \includegraphics[width=\dbimsz]{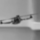} &
      \includegraphics[width=\dbimsz]{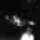} &
      \includegraphics[width=\dbimsz]{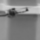} &
      \includegraphics[width=\dbimsz]{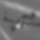} \\
      \includegraphics[width=\dbimsz]{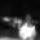} &
      \includegraphics[width=\dbimsz]{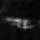} &
      \includegraphics[width=\dbimsz]{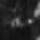} &
      \includegraphics[width=\dbimsz]{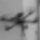} &
      \includegraphics[width=\dbimsz]{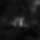} &
      \includegraphics[width=\dbimsz]{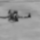} &
      \includegraphics[width=\dbimsz]{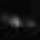} \\
      \includegraphics[width=\dbimsz]{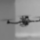} &
      \includegraphics[width=\dbimsz]{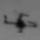} &
      \includegraphics[width=\dbimsz]{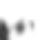} &
      \includegraphics[width=\dbimsz]{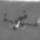} &
      \includegraphics[width=\dbimsz]{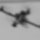} &
      \includegraphics[width=\dbimsz]{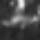} &
      \includegraphics[width=\dbimsz]{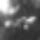} \\
      \includegraphics[width=\dbimsz]{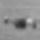} &
      \includegraphics[width=\dbimsz]{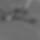} &
      \includegraphics[width=\dbimsz]{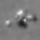} &
      \includegraphics[width=\dbimsz]{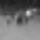} &
      \includegraphics[width=\dbimsz]{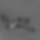} &
      \includegraphics[width=\dbimsz]{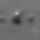} &
      \includegraphics[width=\dbimsz]{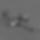} \\
      & & & & & & \\
    \end{tabular} \\
    \hline
  \end{tabular}
  \caption{Sample  real images  both for  training and  evaluation from  the UAV
    dataset. The evaluation  images, while created using a  single UAV, are very
    challenging  as   they  are  low-resolution   while  exhibiting  significant
    lighting, background, and pose variations.}
  \label{fig:datasets}
\end{figure}

\section{Results}
\label{sec:results}

\vincent{In this section,  we first introduce the three datasets that we
  used  for training  and  testing  of our  algorithms.  Then we  
  compare  our  synthetic  data  generation  approach  with several
  baselines, and evaluate the  importance of each  of our
  rendering effects.  Our next
  step is to show  the significance of the optimization  of the $\Theta$
  rendering parameters. Further on we experimentally estimate the
  optimal ratio between synthetic and  real samples used for 
  training. We then show that our  algorithm is able to generalize to
  multiple kinds of aircrafts. Finally we compare our approach to
  a very recent one  on realistic  data generation  on the
  {\bf PASCAL VOC} dataset.}


\begin{itemize}

\item  {\bf UAV  Dataset.} This  dataset contains  challenging images  that were
  acquired from the  camera of a flying UAV. In  these low-resolution images one
  can  see  another  drone  that  flies around  and  appears  against  different
  backgrounds and under various lighting  conditions. Even though only one drone
  was used  to produce the images,  the dataset includes many  of the challenges
  that  outdoor environments  pose, such  as large  illumination  and background
  changes.   We  use it  to  investigate the  impact  of the different effects  our
  rendering pipeline includes.

\item {\bf Aircraft dataset.}  This  dataset contains images of different planes
  seen against changing backgrounds and  under a variety of weather and lighting
  conditions.  We use it to demonstrate  that our approach generalizes to a much
  larger class of objects than simply drones. As in the case of the UAV dataset,
  we will  demonstrate that  regardless of the  machine learning method  used to
  detect  the  target  objects,  we  can improve  performance  by  appropriately
  generating our synthetic images.

\item {\bf PASCAL  VOC 2007.}  We use this  well-known Computer Vision benchmark
  to   compare   our  approach   to   a   very   recent  work   on synthetic   view
  generation~\cite{Rematas14}.   As in~\cite{Rematas14},  we  restrict ourselves
  here to the car class, which nevertheless further demonstrates the versatility
  of our approach.

\end{itemize}

We will  present our results in terms  of both recall  $r$ vs precision $p$  curves and  {\it average  precision} AveP,  defined as  $\int  \limits ^1 \limits _0 p(r)dr$. \artem{Some additional results and video sequences can be found on the webpage of the project\footnote{\url{http://cvlabwww.epfl.ch/~rozantse/synthetic_data.html}}.}

\subsection{Gauging the Various Components of the Approach using the UAV Dataset}

We created  a dataset of 2,000 images  of UAVs in various  environments and seen
under  different lighting conditions.   Fig.~\ref{fig:datasets} depicts  some of
the images.  The images were captured  by one UAV filming another one while they
were both flying.

\begin{figure*}
  \centering
  \begin{tabular}{m{0.1in}ccc}
    \vspace{-1in} \rotatebox{90}{Real Data} & 
    \includegraphics[width=1.2in]{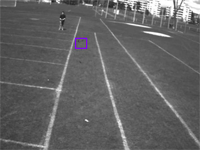}  &
    \includegraphics[width=1.2in]{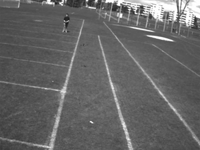} &
    \includegraphics[width=1.2in]{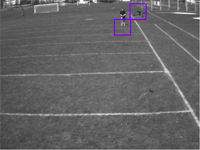} \\ 
    \vspace{-1in} \rotatebox{90}{Real+Synth Data} & 
    \includegraphics[width=1.2in]{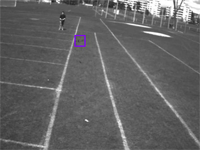} &
    \includegraphics[width=1.2in]{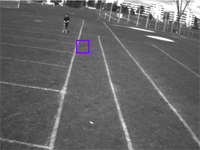}  &
    \includegraphics[width=1.2in]{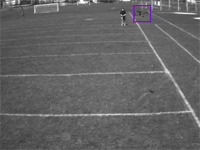}  \\ 
    \\
    \vspace{-1in} \rotatebox{90}{Real Data} & 
    \includegraphics[width=1.2in]{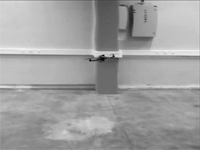} &
    \includegraphics[width=1.2in]{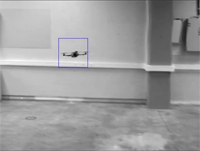}  &
    \includegraphics[width=1.2in]{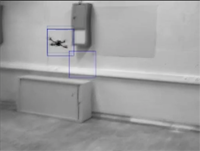}  \\ 
    \vspace{-1in} \rotatebox{90}{Real+Synth Data} & 
    \includegraphics[width=1.2in]{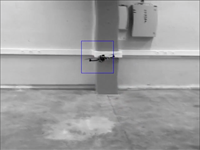} &
    \includegraphics[width=1.2in]{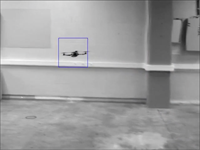}  &
    \includegraphics[width=1.2in]{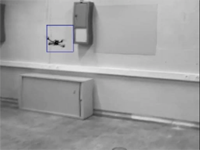}  \\ 
  \end{tabular}
  \caption{ \pascal{Qualitative  comparison of the performance  of the detectors
      trained just on real data versus both real and synthetic data.  }}
  \label{fig:comp_Detections_UAV}
\end{figure*}

\begin{figure*}
  \centering
  \begin{tabular}{ccc}
  \multicolumn{3}{c}{Detections} \\
    \includegraphics[width=\threeim]{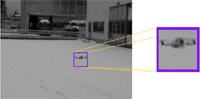} &
    \includegraphics[width=\threeim]{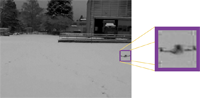}  &
    \includegraphics[width=\threeim]{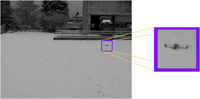}  \\ 
    \includegraphics[width=\threeim]{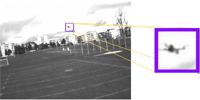}  &
    \includegraphics[width=\threeim]{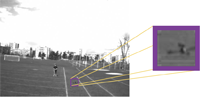}  &
    \includegraphics[width=\threeim]{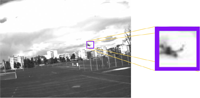}  \\ 
    \includegraphics[width=\threeim]{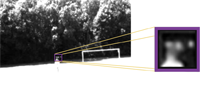}  &
    \includegraphics[width=\threeim]{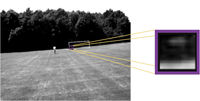}  &
    \includegraphics[width=\threeim]{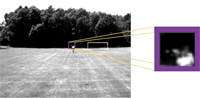}  \\
    \multicolumn{3}{c}{Missed and False Detections} \\
    \includegraphics[width=\threeim]{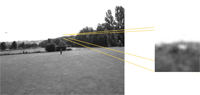} &
    \includegraphics[width=\threeim]{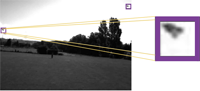}  &
    \includegraphics[width=\threeim]{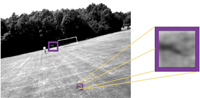}  \\
    
  \end{tabular}
  \caption{{\bf  Top  rows:}  More  sample detections  by  the  AdaBoost
      detector trained  using both real  and synthetic data.  {\bf  Bottom row:}
      Missed or  false detections to  highlight the challenges that  the dataset
      provides for the detector. (best seen in color)}
  \label{fig:Detections_UAV}
\end{figure*}

Fig.~\ref{fig:comp_Detections_UAV}  depicts  detections  by an  AdaBoost
classifier trained  using either real images  only or both  real and synthetic
images.   We  will  quantify  the  observed  performance  improvement  in  the
remainder  of   this  section.   In   Fig.~\ref{fig:Detections_UAV},  we  show
additional examples  of detections  by the detector  trained on both  real and
synthetic data  as well  as some failure  cases to illustrate  how challenging
this dataset is.

We first describe the acquisition process  and then use these UAV images to test
individual components of our pipeline and to evaluate overall performance. 

\subsubsection{Experimental Setup}
\label{subsec:exp_setup}

To obtain the background images required  to render the composite ones, we first
aligned consecutive frames by computing the homographies between the frames, and
kept the median intensity at each location of the aligned images.

The  training and  testing videos  were acquired  in different  environments and
feature different backgrounds.  The CAD model of the UAV used for rendering only
coarsely  outlines  the  main  geometrical  structure of  the  real  object,  as
illustrated by Fig.~\ref{fig:im_synth}(a). Negative training and testing samples
were obtained by randomly sampling the  backgrounds of the training images.  For
detection, we use  a sliding window approach that applies  the detector at every
spatial  location and  at  different scales  of the  whole  image.  Non-maximum
suppression is then applied to the response image scale-space.

The detection methods in the experiments  are trained with a combination of
real and synthetic data and tested on the real data only.

\subsubsection{Comparing against simply Perturbing the Real Images}

A broadly used approach to augmenting a training set is to perturb the available
images     using     simple    image     transformations~\cite{LeCun98,Varga03}.
Table~\ref{tab:comp_with_baseline}  compares  the   performances  of  all  three
selected  detectors when being  trained on  images generated either in this  way or
using our approach.  The  perturbations involve combining rotation, translation,
mirroring, blurring and adding noise to the original images.

Our  approach  significantly outperforms  this  simple  technique. This  can  be
explained by  the fact that we  generate realistic combinations of  3D poses and
background that are not present in the seed images.

\begin{table*}
  \centering
  \ra{1}
  \begin{tabular}{@{}lcccc@{}}
    \toprule
    & \phantom{abcabcabcabcabc} & \scriptsize{Using real} & \scriptsize{By perturbing} & \multirow{2}{*}[0.4em]{Our method} \\[-0.7em]
    & &\scriptsize{images only}& \scriptsize{the real images} & \\
    \midrule
    \scriptsize{Detection method:}&&\multicolumn{3}{c}{\scriptsize{Average precision:}}\\
    DPM      && 0.84 & 0.87 & \bf{0.93} \\
    AdaBoost && 0.80 & 0.83 & \bf{0.92} \\
    CNN      && 0.85 & 0.86 & \bf{0.89} \\
    \bottomrule
  \end{tabular} 
  \caption{\label{tab:comp_with_baseline} Comparing  average precisions for each
    detection method  when either perturbing  real training images or  using our
    approach with  the optimal  number of synthetic  images and  the appropriate
    distance measure .  Our  approach significantly outperforms this traditional
    method.}
\end{table*}

\subsubsection{Relative Importance of the Various Rendering Effects}
\label{subsec:eval}

\begin{figure*}
  \centering
  \begin{tabular}{c}
    \begin{tabular}{ccc}
      \includegraphics[width=\threeim]{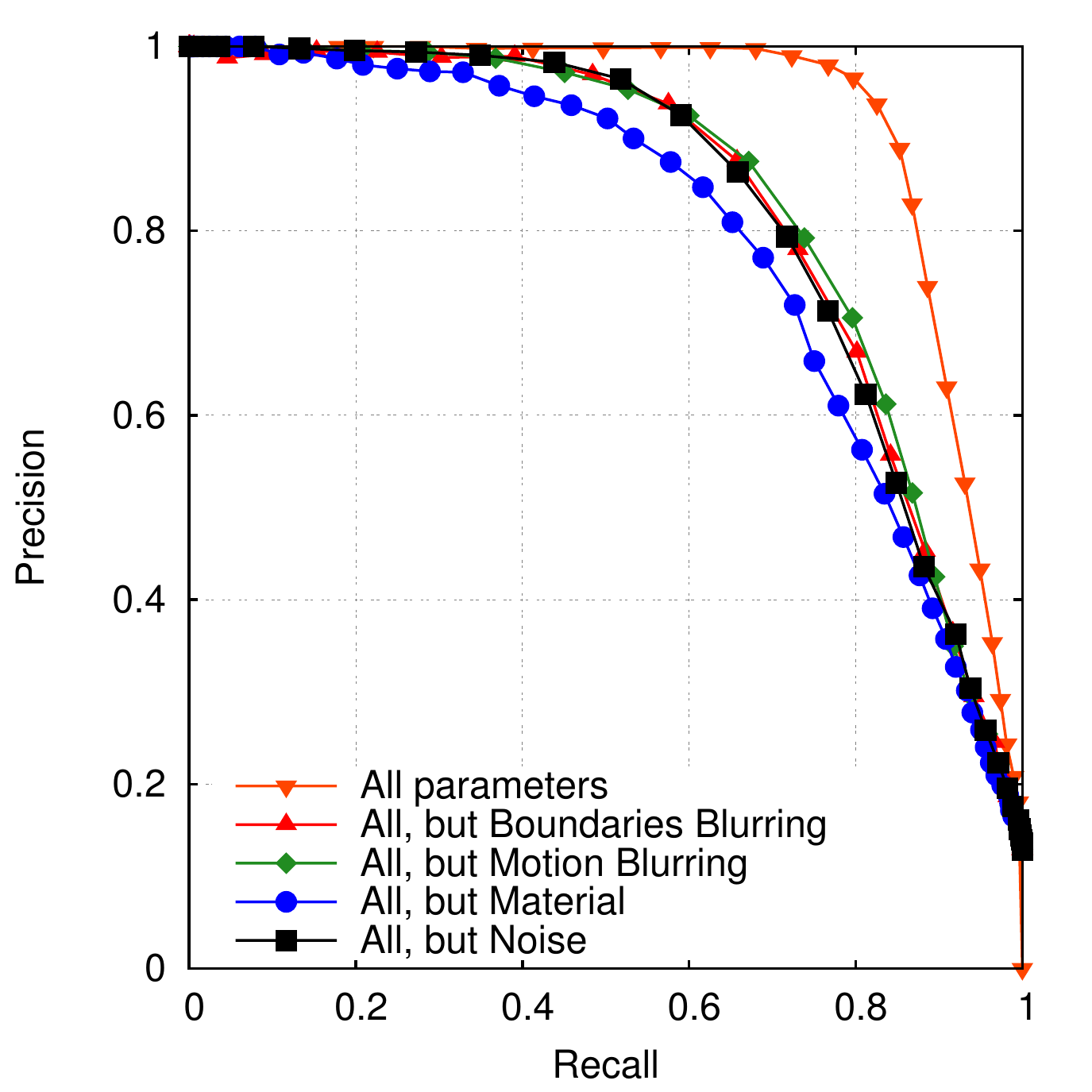} &
      \includegraphics[width=\threeim]{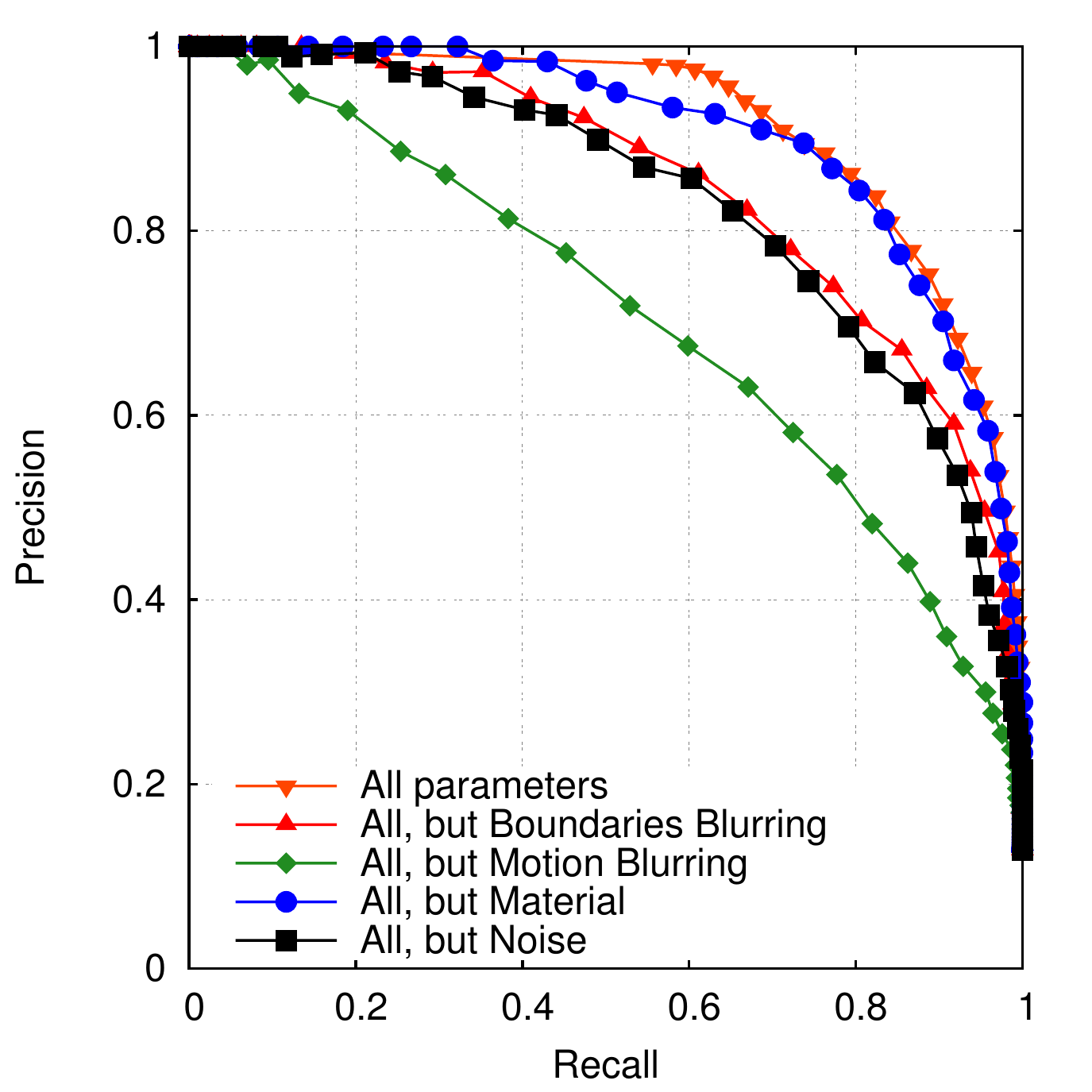} &
      \includegraphics[width=\threeim]{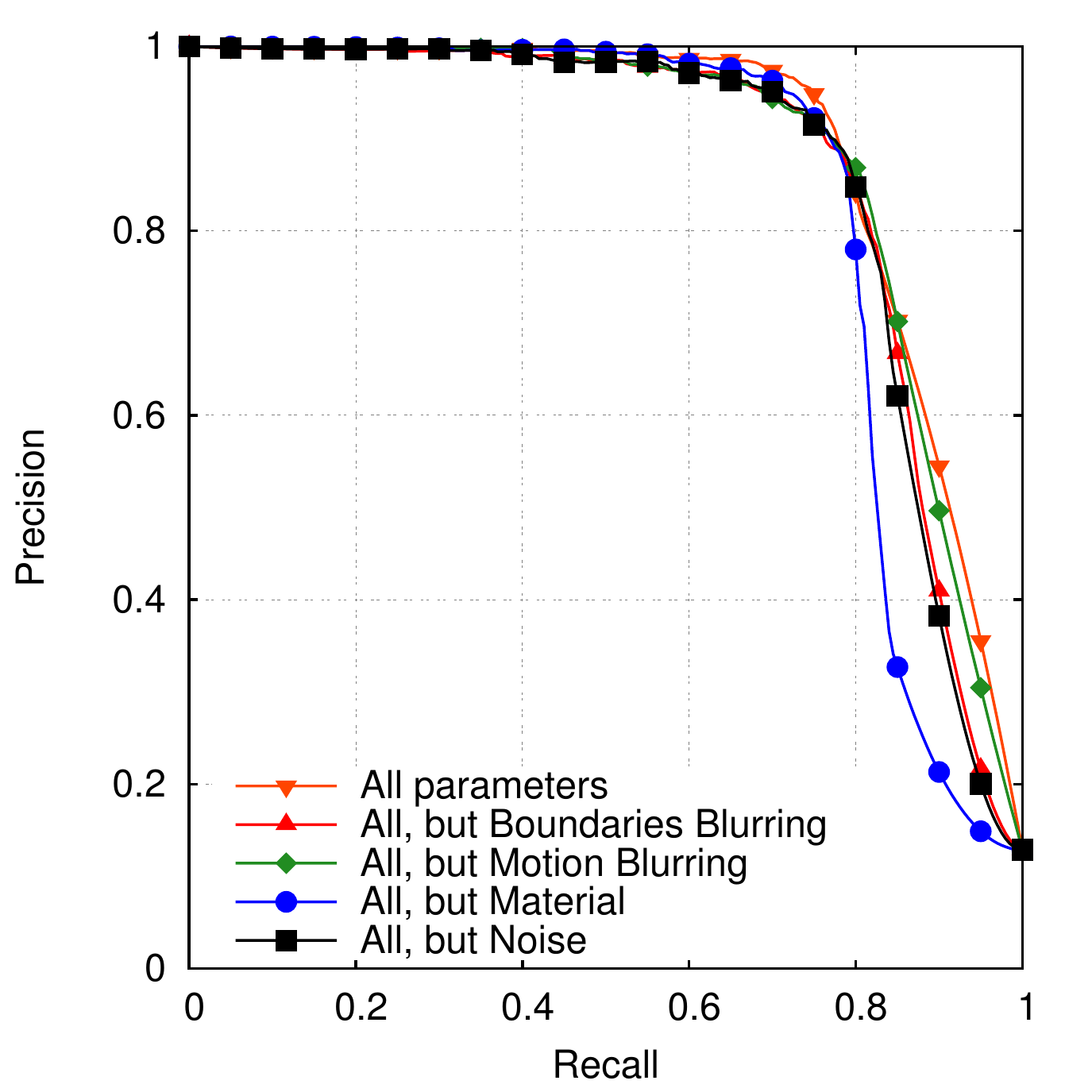} \\
      DPM and $d_\HOG(.,.)$ & AdaBoost and $d_\WL^L(.,.)$  & CNN and $d_\CNN(.,.)$ \\
    \end{tabular}\\
    \\
    \ra{1}
    \begin{tabular}{@{}lcc@{\hspace{0.4cm}}c@{\hspace{0.4cm}}c@{\hspace{0.4cm}}c@{\hspace{0.4cm}}c@{}}
      \toprule
      & \phantom{abcabcabc} & \multicolumn{5}{c}{\scriptsize{Synthetic data generation effects:}} \\
      \cmidrule{3-7}
      &             & All & no \bf{BB} & no \bf{MB} & no \bf{RN}  & no \bf{MP} \\
      \midrule
      \scriptsize{Detection method:}& & \multicolumn{5}{c}{\scriptsize{Average precision:}}\\
      DPM      && \bf{0.93} & 0.83 & 0.84 & 0.83 & 0.80\\
      AdaBoost && \bf{0.92} & 0.85 & 0.71 & 0.75 & 0.91 \\
      CNN      && \bf{0.89} & 0.88 & \bf{0.89} & 0.88 & 0.85 \\
      \bottomrule
    \end{tabular} 

  \end{tabular} 
  \caption{Evaluating  the importance  of each  \intrinsic parameter.   Each one
    clearly  has a  positive influence  of the  quality of  the  synthetic data.
    However, their respective impacts  depends on the specific detection method.
    (best seen in color)}
  \label{fig:eff_eval}
\end{figure*}

To  demonstrate that  correctly setting  each one  of the  \intrinsic parameters
introduced  in  Section~\ref{sec:synth_data}  truly  matters, we  performed  the
following set of experiments. For each effect---object boundary blurring, Motion
blurring, Random noise, Material  properties---we set the corresponding value in
the \intrinsic  parameters $\Theta$ of  Eq.~\ref{eq:Intrinsic} to 0  to suppress
its  influence  and  optimized   the  other  parameters  using  the  appropriate
similarity  measure for  each  detection  method.  We  then  used the  resulting
$\Theta$'s  to   generate  the  synthetic  images,   trained  the  corresponding
detector on these images and evaluated is on the test images.  The results are
shown in  Fig.~\ref{fig:eff_eval}. Correctly modeling each effect  clearly has a
positive influence on final performance.

\subsubsection{Importance of Optimizing over the Rendering Parameters}

To show the importance of optimizing over the \intrinsic parameters $\Theta$, we
compare  in   Fig.~\ref{fig:rand_par}  the  final   performance  obtained  using
optimized parameters with the  final performance obtained with random parameters
drawn from a uniform distribution. The minimum and maximum values of the uniform
distribution  were taken  as the  minimum and  maximum values  of  the optimized
parameters.   Our  optimization-based  approach  clearly  brings  a  significant
improvement.

\newcommand{\simsz}{1.4in}

\begin{figure}
\centering
\begin{tabular}{c}
\begin{tabular}{ccc}
\includegraphics[width=\simsz]{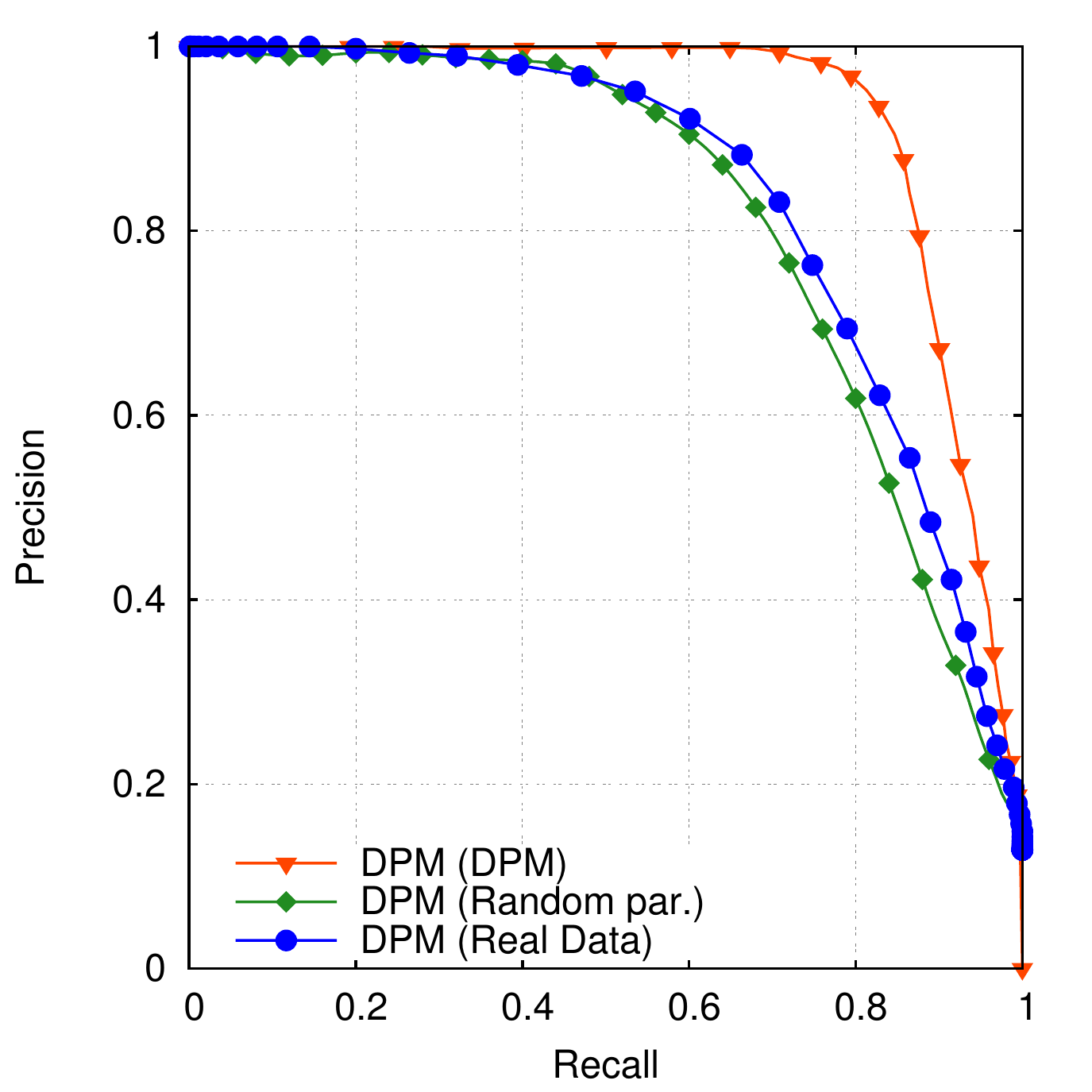} &
\includegraphics[width=\simsz]{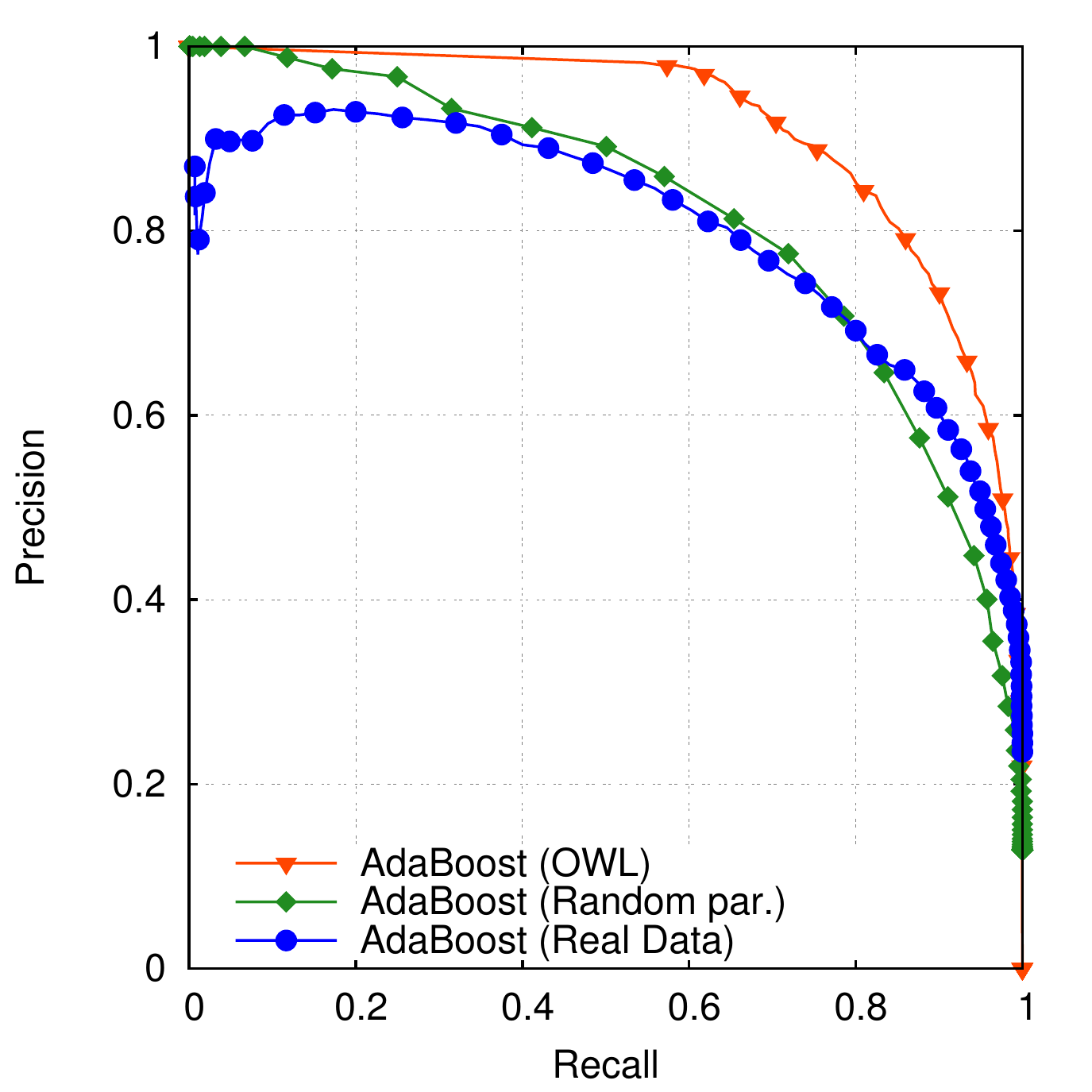} &
\includegraphics[width=\simsz]{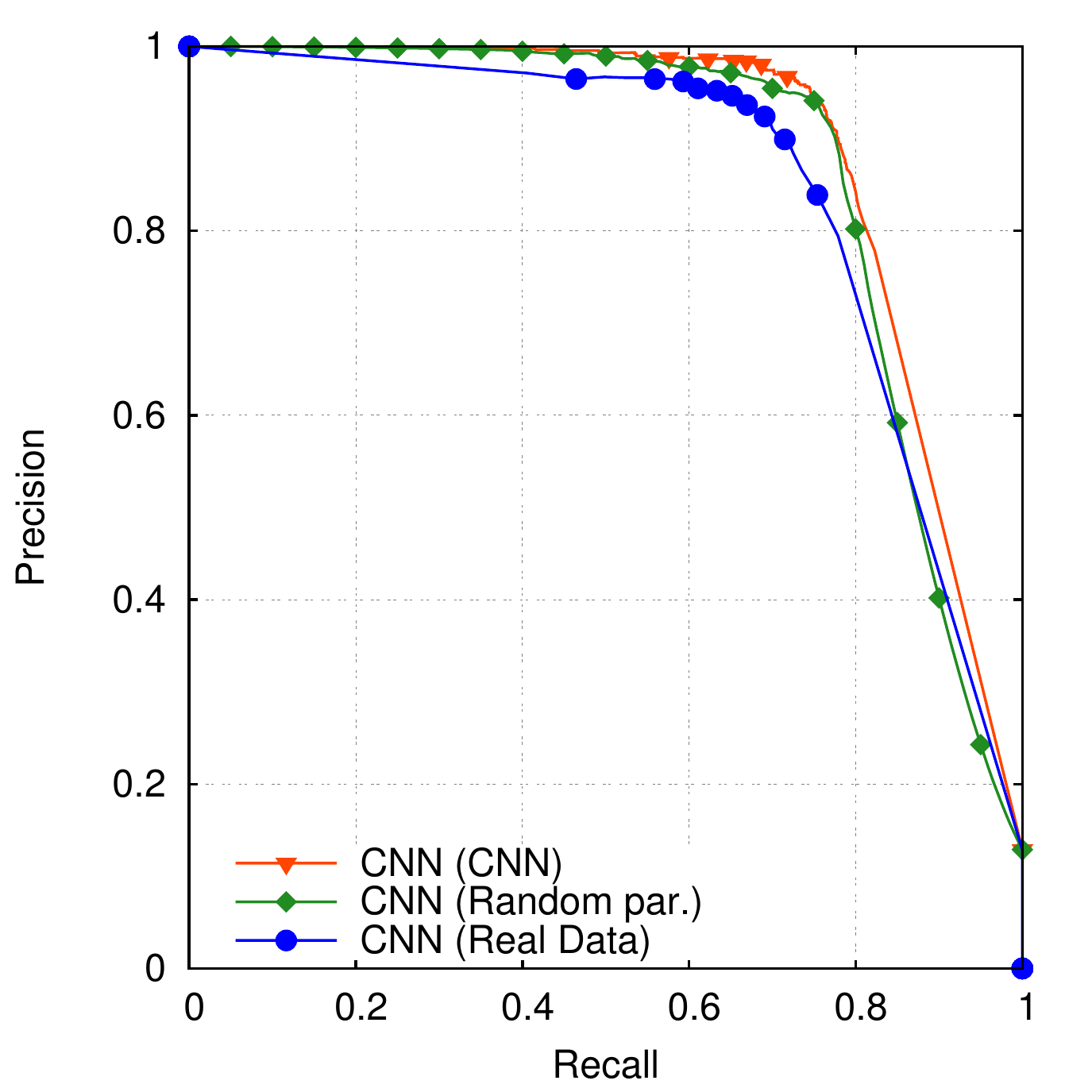} \\
DPM & AdaBoost & CNN
\end{tabular} \\
\\
\begin{tabular}{@{}lc@{\hspace{4cm}}c@{\hspace{1cm}}c@{\hspace{1cm}}c@{}}
\toprule
&               & \scriptsize{Using real}    & \scriptsize{Random} & \scriptsize{Optimized} \\[-0.7em]
&               & \scriptsize{images only} & \scriptsize{parameters} & \scriptsize{parameters} \\
\midrule
\scriptsize{Classification method:}&&\multicolumn{3}{c}{\scriptsize{Average precision:}}\\
\cmidrule{3-5}
DPM      && 0.84 & 0.82 & \bf{0.93} \\
AdaBoost && 0.80 & 0.82 & \bf{0.92}  \\
CNN      && 0.85 & 0.87 & \bf{0.89}  \\
\bottomrule
\end{tabular} 

\end{tabular}
\caption{Comparison of the performances of different detectors trained on real
  and synthetic data generated using corresponding similarity measures with
  those where the \intrinsic parameters are randomly selected. The optimized
  parameters always yield better performance. (best seen in color)}
\label{fig:rand_par}
\end{figure}

\subsubsection{Influence of the Number of Synthetic and Real Images}

To  evaluate how much  we can  improve the  performances using  synthetic images
generated  with  our approach,  we  trained each  of  the  detection methods  we
consider with different numbers of synthetic samples in addition to the real training
samples.  For each detector, the synthetic samples were generated using the
parameters obtained using the appropriate similarity functions.

\begin{figure}
  \centering
  \begin{tabular}{cc}
    \includegraphics[width=2.1in]{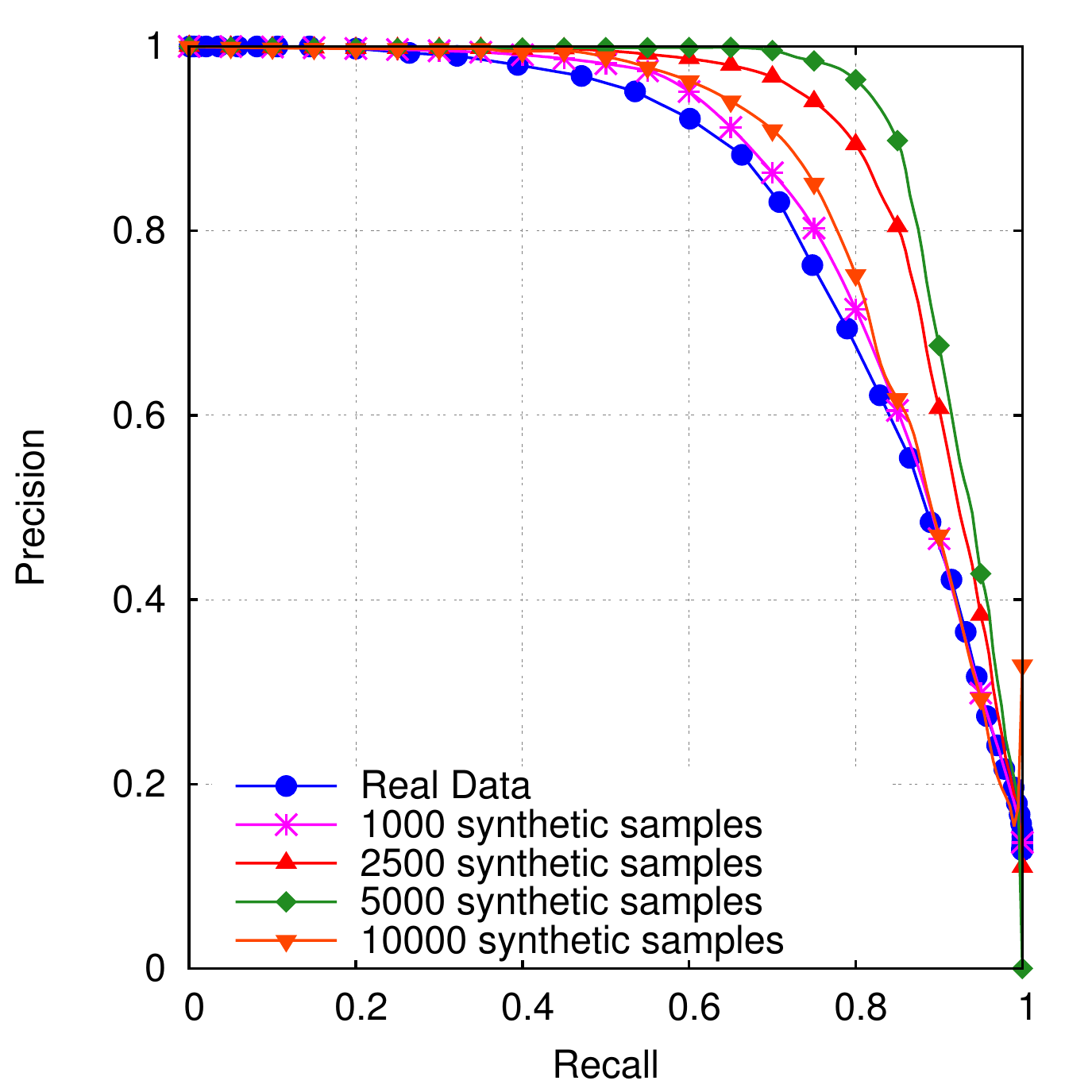} & 
    \includegraphics[width=2.1in]{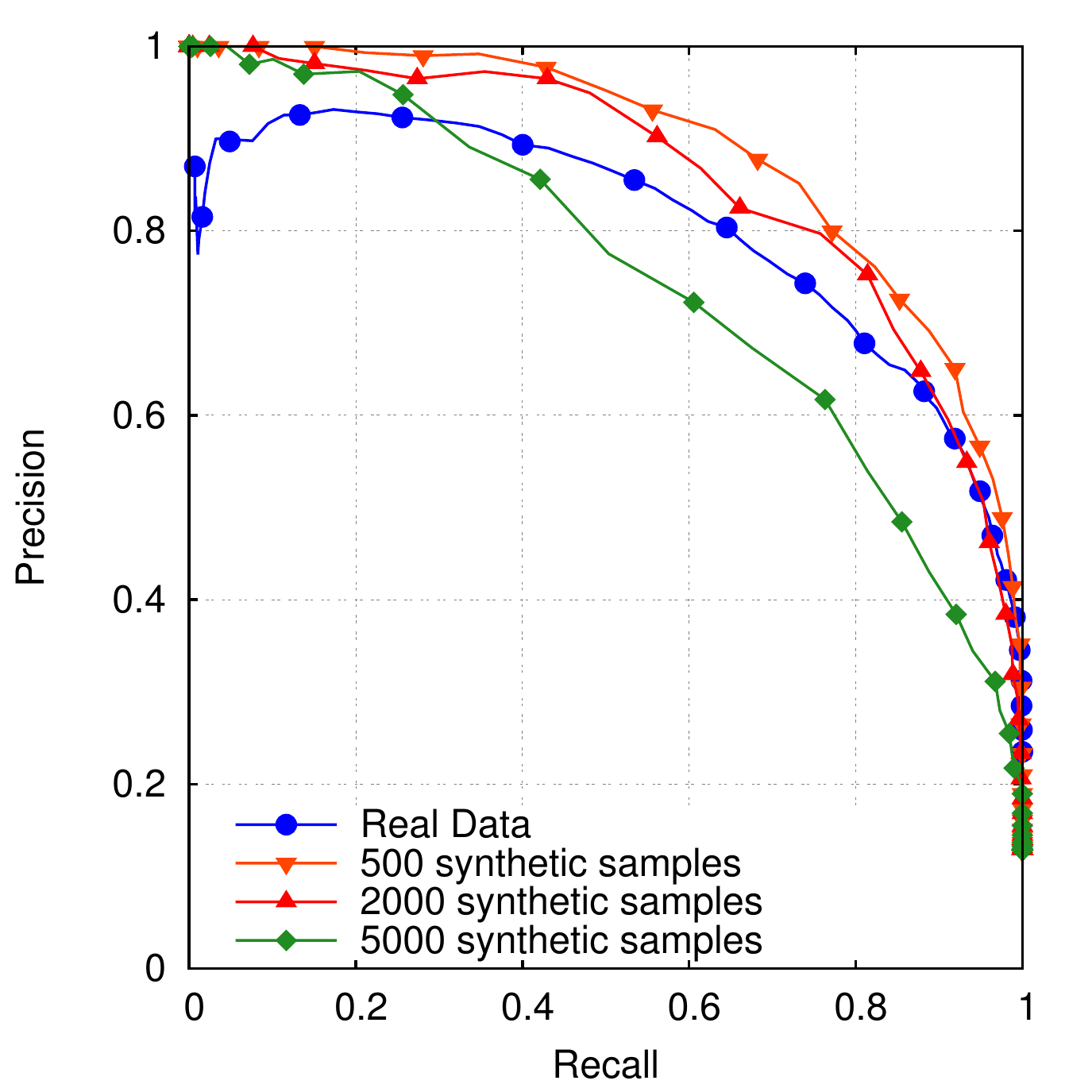} 
    \\
    DPM and $d_\HOG(.,.)$ &     AdaBoost and $d_\WL^R(.,.)$ \\
    \includegraphics[width=2.1in]{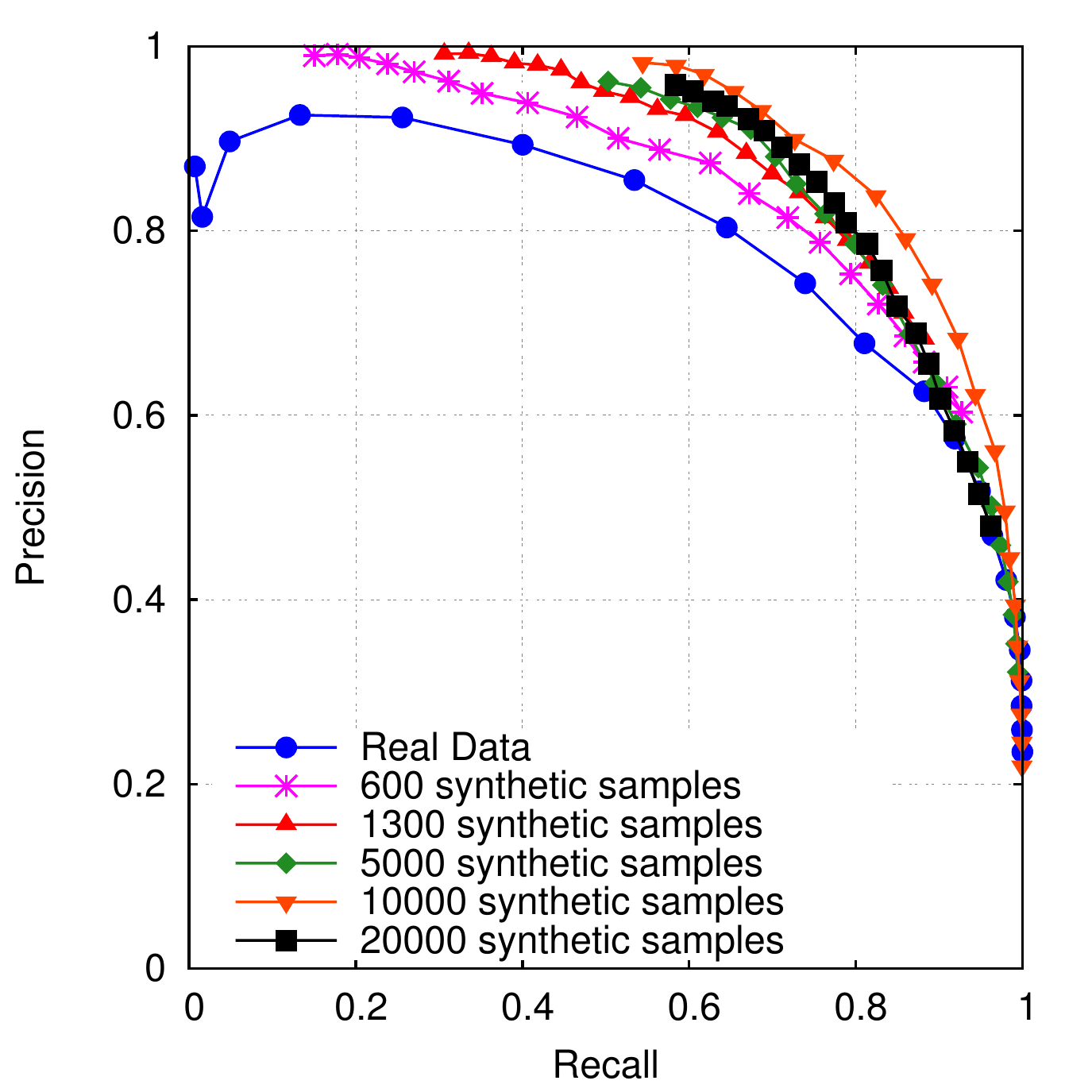} &
    \includegraphics[width=2.1in]{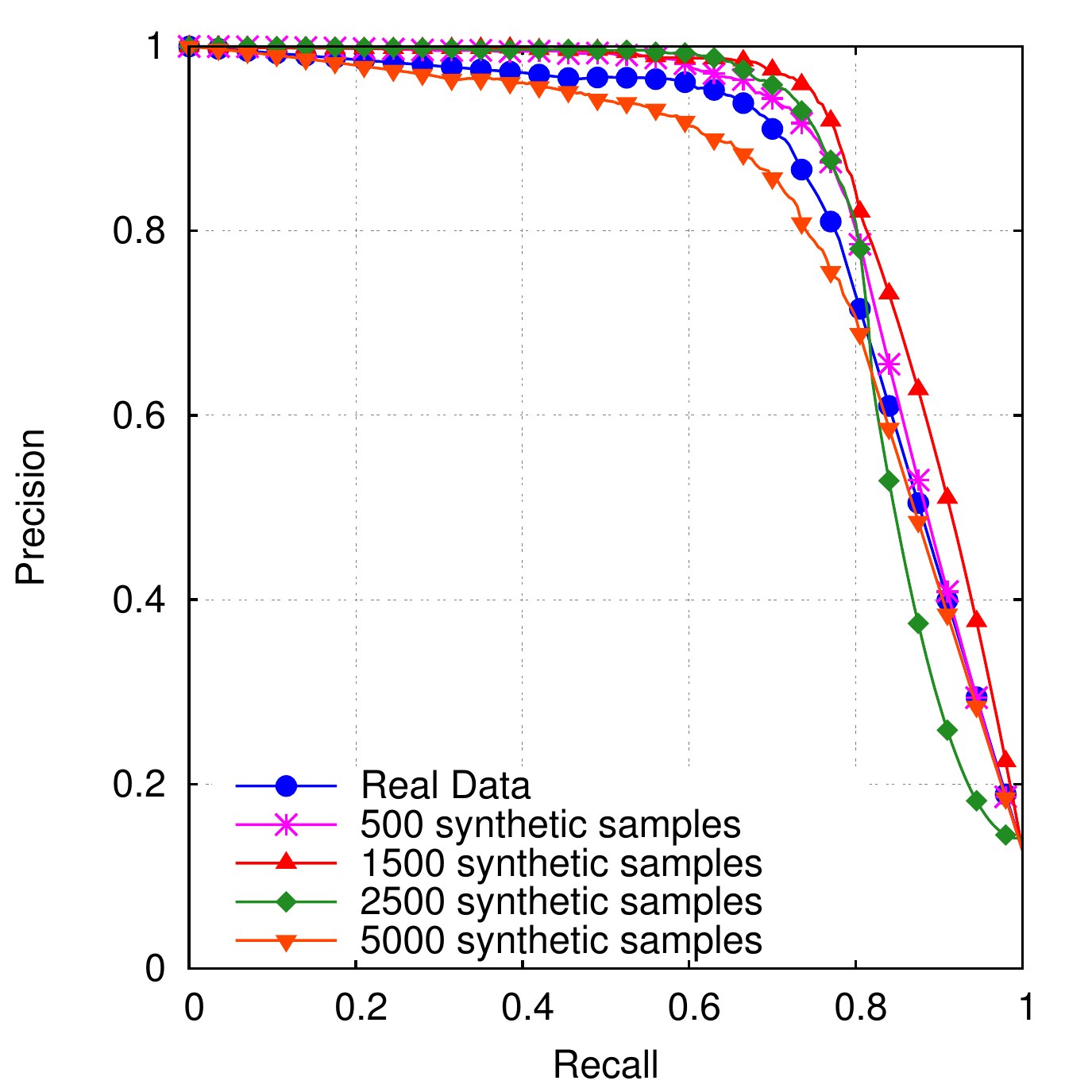} \\
    AdaBoost and $d_\WL^L(.,.)$ & CNN and $d_\CNN(.,.)$\\ 
  \end{tabular}
  \caption{We varied the number of synthetic images used for training, for three
    detection  methods, using  their corresponding  similarity  measures.  Using
    both  real and  synthetic  data for  training  phase increases  performances
    compared to real data used alone.  However using too much synthetic data may
    also hurt. (best seen in color)}
  \label{fig:UAV_eff}
\end{figure}

Fig.~\ref{fig:UAV_eff}  compares the  performances of  these detectors when varying the  number of synthetic samples.  It can be  seen that using the synthetic images significantly improves  performance over using the real images alone. However,  this is only true  up to a  point.  When there are  too many
synthetic  images,  the performance  eventually  {\it  decreases} because  the influence of the  real images gets drowned out.  In  practice, this means that for best performance, it makes sense  to use a validation set to ascertain the  optimal ratio of synthetic to real images.

\vincent{  From  these experiments  we  can  conclude  that  the best  ratio  of
  synthetic and  real examples that should  be used for training  depends on the
  detection  algorithm.   AdaBoost  achieves  its highest  accuracy  with  $100$
  synthetic images for each real one, DPM with 50 synthetic images for each real
  one, and CNN with $15-20$ synthetic images for each real one. }

\begin{figure}
  \centering
  \begin{tabular}{ccc}
    \includegraphics[width=\threeim]{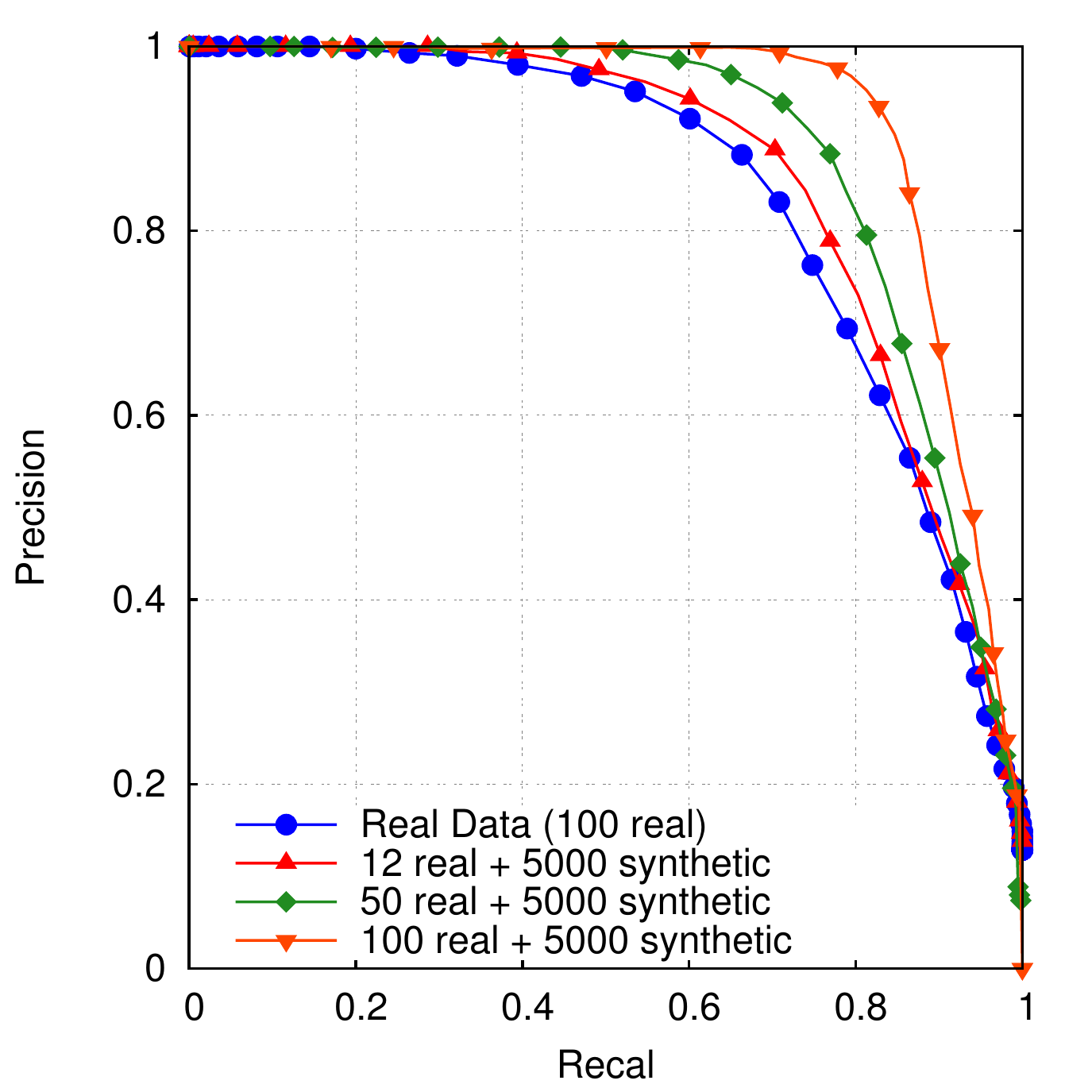} &
    \includegraphics[width=\threeim]{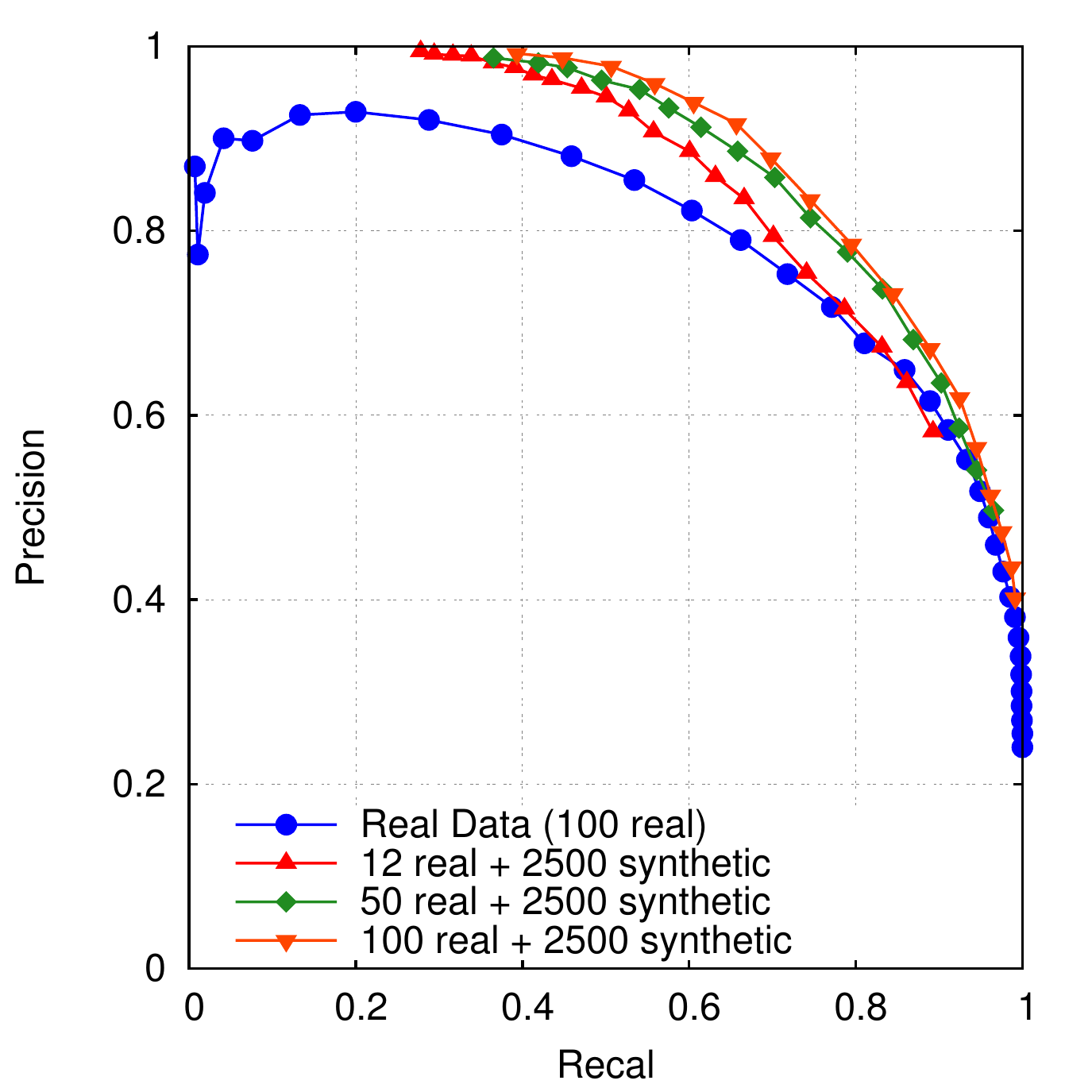} &
    \includegraphics[width=\threeim]{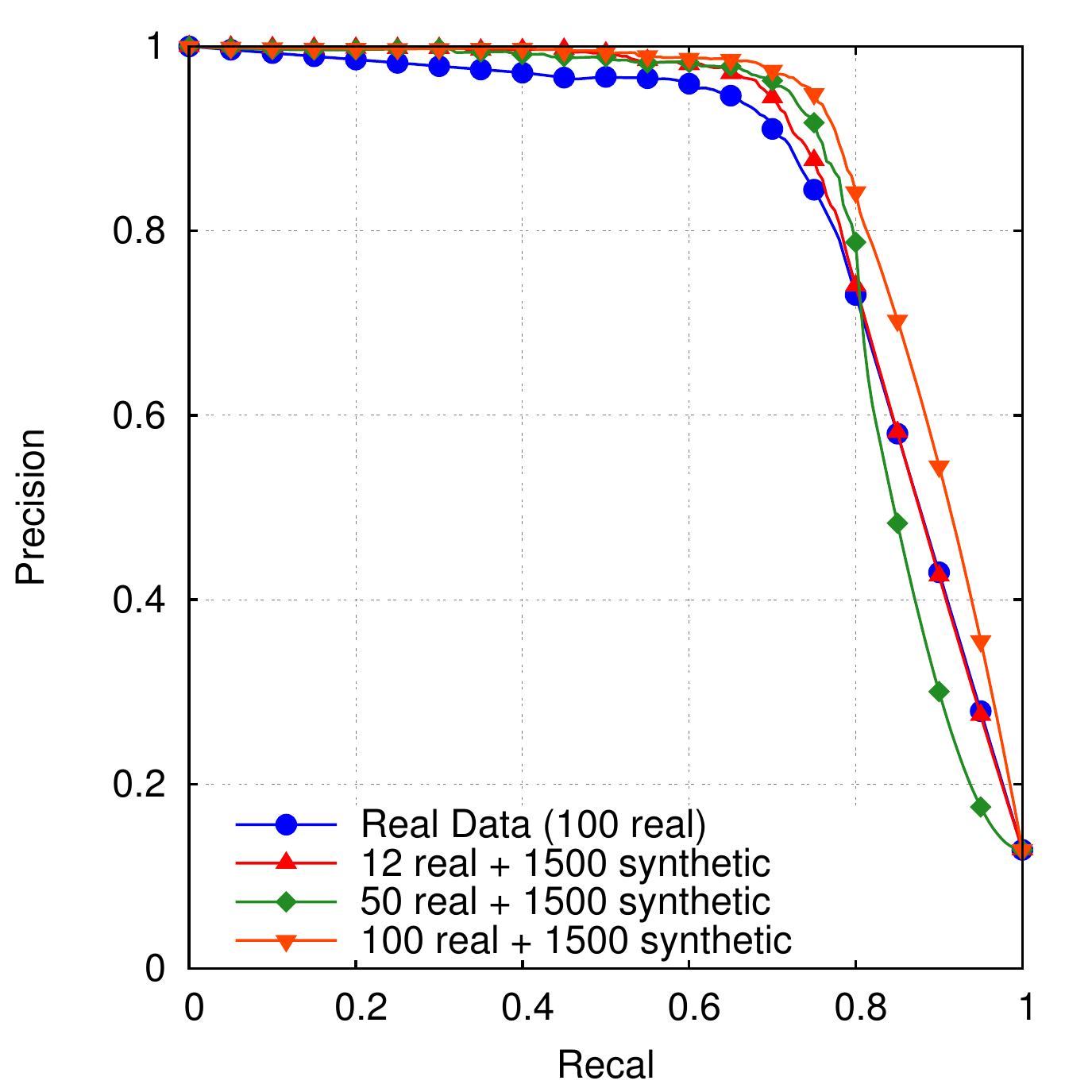} \\
    DPM and $d_\HOG(.,.)$ & AdaBoost and $d_\WL^L(.,.)$  & CNN and $d_\CNN(.,.)$ \\
  \end{tabular}
  \caption{Performance of  the DPM, AdaBoost, and CNN  detectors for different
    numbers of seed  real images. Using as  few as 12 real samples  is enough to
    generate synthetic samples that allow  us to outperform a detector trained
    with 100 images.   For each detector, the synthetic  images were generated
    using the corresponding similarity measure.}
  \label{fig:UAV_rim_eff}
\end{figure}

We also evaluated the  influence of the number of seed real  images on the final
performances,  by decreasing  the number  of real  images used  to  optimize the
rendering  parameters.   Fig.~\ref{fig:UAV_rim_eff} shows  the  results for  the
AdaBoost detector.   Using as  few as  12 real samples  is enough  to generate
synthetic samples that allows us to outperform a detector trained with about 8
times as  many real images. Unsurprisingly,  increasing the number  of seed real
images results in an improvement of the final performances.

\begin{table*}
  \centering
  \ra{1}
  \begin{tabular}{@{}lcccccc@{}}
    \toprule
    & &\multicolumn{5}{c}{\scriptsize{Similarity measure:}} \\
    \cmidrule{3-7}
    & \scriptsize{Using real} & \multirow{2}{*}[0.4em]{$d_\Eucl(.,.)$} &
    \multirow{2}{*}[0.4em]{$d_\HOG(.,.)$} & \multirow{2}{*}[0.4em]{$d_\WL^R(.,.)$} & \multirow{2}{*}[0.4em]{$d_\WL^L(.,.)$} & \multirow{2}{*}[0.4em]{$d_\CNN(.,.)$}  \\[-0.7em]
    &\scriptsize{images only}&&&&&\\
    \midrule
    \scriptsize{Detection method:}&\multicolumn{6}{c}{\scriptsize{Average precision:}}\\
    DPM      & 0.84 & 0.78 & \bf{0.93} & 0.70 & 0.72      & 0.67 \\
    AdaBoost & 0.80 & 0.72 & 0.85      & 0.89 & \bf{0.92} & 0.75 \\
    CNN      & 0.85 & 0.84 & 0.84      & 0.84 & 0.86      & \bf{0.89} \\
    \bottomrule
  \end{tabular} 
  \caption{\label{tab:comp}   Comparison   of    average   precisions   for   each
    detection method,  when the optimal  number of synthetic images  is used.
    Each  detection  method performs  best  with  the corresponding  similarity
    function.}
\end{table*}

\subsubsection{Optimal Performance}
\label{subsec:comp_stoa}

In  this section,  for each  detection  method, we  use the  optimal numbers  of
synthetic samples as discussed in the previous section. For comparison purposes,
we  also estimated  the  optimal numbers  of  synthetic samples  when using  the
Euclidean distance $d_\Eucl(.,.)$ as similarity measure.

Table~\ref{tab:comp} confirms that each  detection method performs best when trained using synthetic images, generated using appropriate similarity measure,       as discussed in Section~\ref{subsec:sim_measure}. In particular, using the Euclidean distance is not only ineffective, but actually yields worse results than not using synthetic images at all.  Interestingly, the best performance is obtained with DPM trained with both real  and synthetic  images, even though  CNN was  better than DPM  when no synthetic images were used.

\subsection{Detecting Multiple Kinds of Aircrafts}

\begin{figure*}
  \centering
  \begin{tabular}{ccc}
    \includegraphics[width=1.4in]{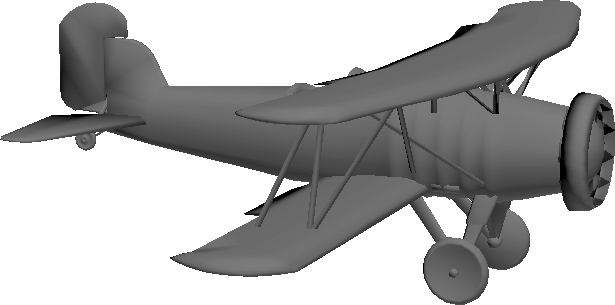} &
    \includegraphics[width=1.4in]{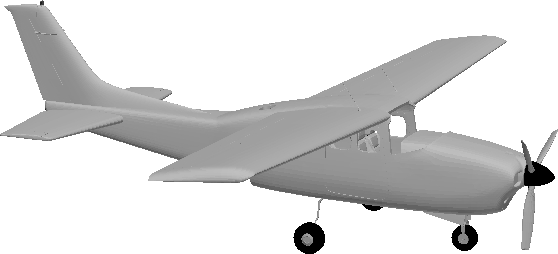}  &
    \includegraphics[width=1.4in]{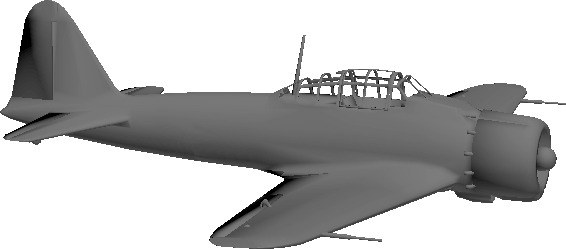}  \\
  \end{tabular}
  \caption{The  three CAD  models we  used for  the Aircraft  dataset.  They are
    freely available on the internet.}
  \label{fig:aircraft_models}
\end{figure*}

For the Aircraft dataset, we  generated synthetic data using CAD models depicted
by  Fig.~\ref{fig:aircraft_models} of  three types  of fixed-wing  aircrafts and
tested them on  different real video sequences. We use 100  real images of these
three aircraft  types along with  their corresponding background  images.  These
images were collected by manually annotating different video sequences where the
aircrafts fly  in different weather  conditions and appear at  different angles.
Sample images from this dataset are shown on Fig.~\ref{fig:aircraft_seed}.

\newcommand{\asize}{0.52in}

\begin{figure*}
  \centering
  \begin{tabular}{ccccccc}
    \includegraphics[width=\asize]{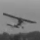}  &
    \includegraphics[width=\asize]{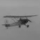}  &
    \includegraphics[width=\asize]{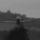}  &
    \includegraphics[width=\asize]{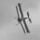}  &
    \includegraphics[width=\asize]{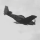}  &
    \includegraphics[width=\asize]{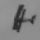}  &
    \includegraphics[width=\asize]{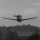}  \\
    \includegraphics[width=\asize]{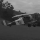}  &
    \includegraphics[width=\asize]{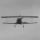}  &
    \includegraphics[width=\asize]{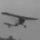}  &
    \includegraphics[width=\asize]{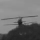}  &
    \includegraphics[width=\asize]{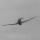}  &
    \includegraphics[width=\asize]{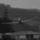}  &
    \includegraphics[width=\asize]{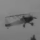}  \\
    \includegraphics[width=\asize]{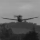}  &
    \includegraphics[width=\asize]{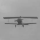}  &
    \includegraphics[width=\asize]{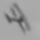}  &
    \includegraphics[width=\asize]{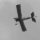}  &
    \includegraphics[width=\asize]{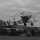}  &
    \includegraphics[width=\asize]{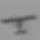}  &
    \includegraphics[width=\asize]{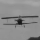}  \\
  \end{tabular}
  \caption{Sample seed real images from the Aircraft dataset.}
  \label{fig:aircraft_seed}
\end{figure*}

\begin{figure*}
  \centering
  \begin{tabular}{ccccccc}
    \includegraphics[width=\asize]{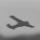} &
    \includegraphics[width=\asize]{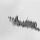}  &
    \includegraphics[width=\asize]{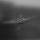}  &
    \includegraphics[width=\asize]{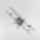}  &
    \includegraphics[width=\asize]{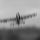}  &
    \includegraphics[width=\asize]{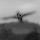}  &
    \includegraphics[width=\asize]{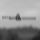}  \\
    \includegraphics[width=\asize]{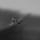}  &
    \includegraphics[width=\asize]{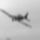}  &
    \includegraphics[width=\asize]{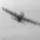}  &
    \includegraphics[width=\asize]{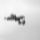}  &
    \includegraphics[width=\asize]{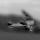}  &
    \includegraphics[width=\asize]{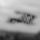}  &
    \includegraphics[width=\asize]{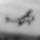}  \\
    \includegraphics[width=\asize]{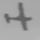}  &
    \includegraphics[width=\asize]{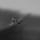}  &
    \includegraphics[width=\asize]{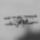}  &
    \includegraphics[width=\asize]{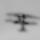}  &
    \includegraphics[width=\asize]{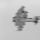}  &
    \includegraphics[width=\asize]{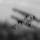}  &
    \includegraphics[width=\asize]{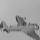}  \\
  \end{tabular}
  \caption{Generated samples of aircrafts.}
  \label{fig:aircraft_synth_im}
\end{figure*}

Here, we  used an  AdaBoost detector trained  using real and  synthetic images
generated    based    on    the    $d_\WL^L(.,.)$   similarity    function    of
Section~\ref{subsec:sim_measure}.   We  generated  10,000 synthetic  samples  to
supplement   the  real   images   and   used  them   as   a  training   dataset.
Fig.~\ref{fig:aircraft_synth_im} depicts sample synthetic images.

\begin{figure*}
  \centering
  \begin{tabular}{ccc}
    \includegraphics[width=\threeim]{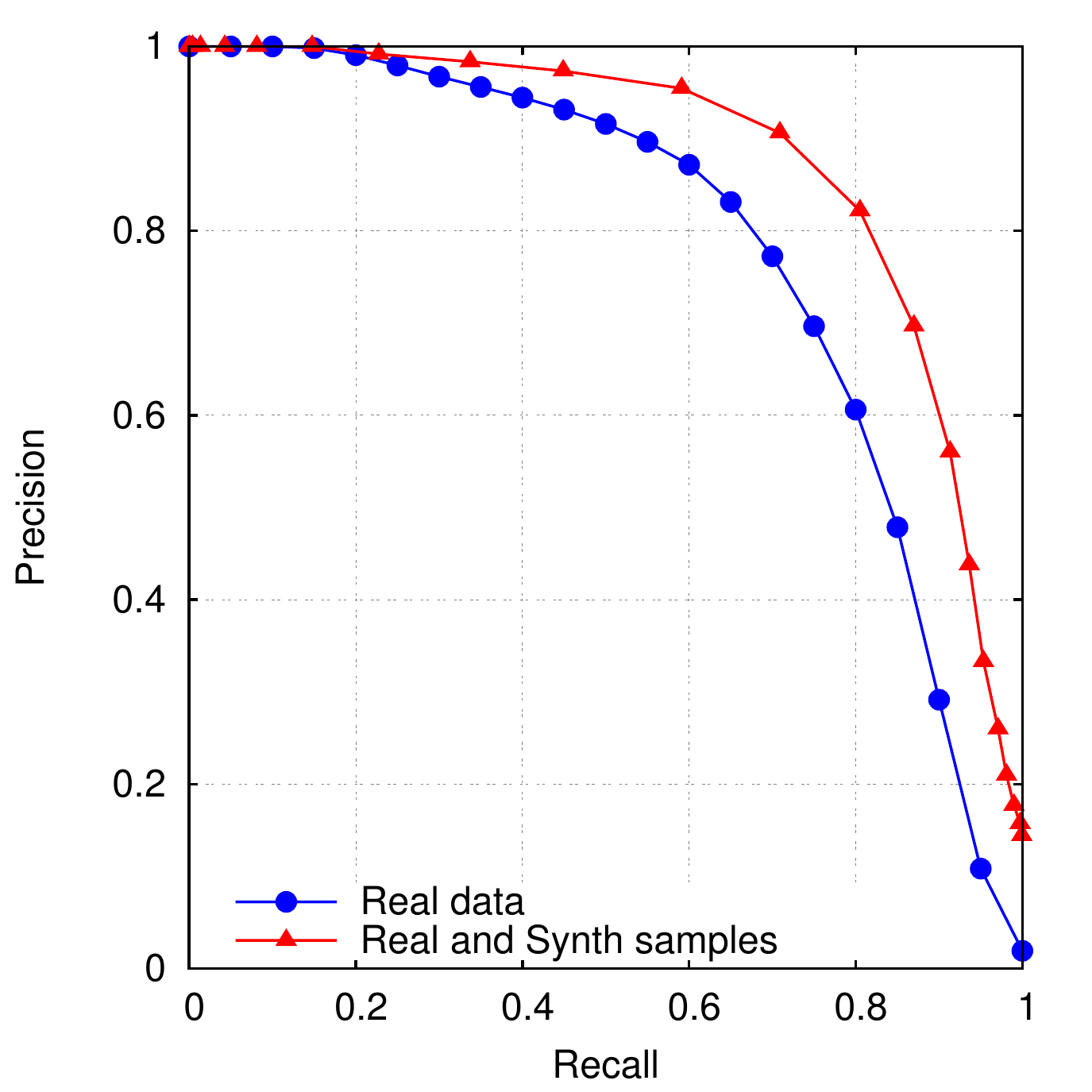} &
    \includegraphics[width=\threeim]{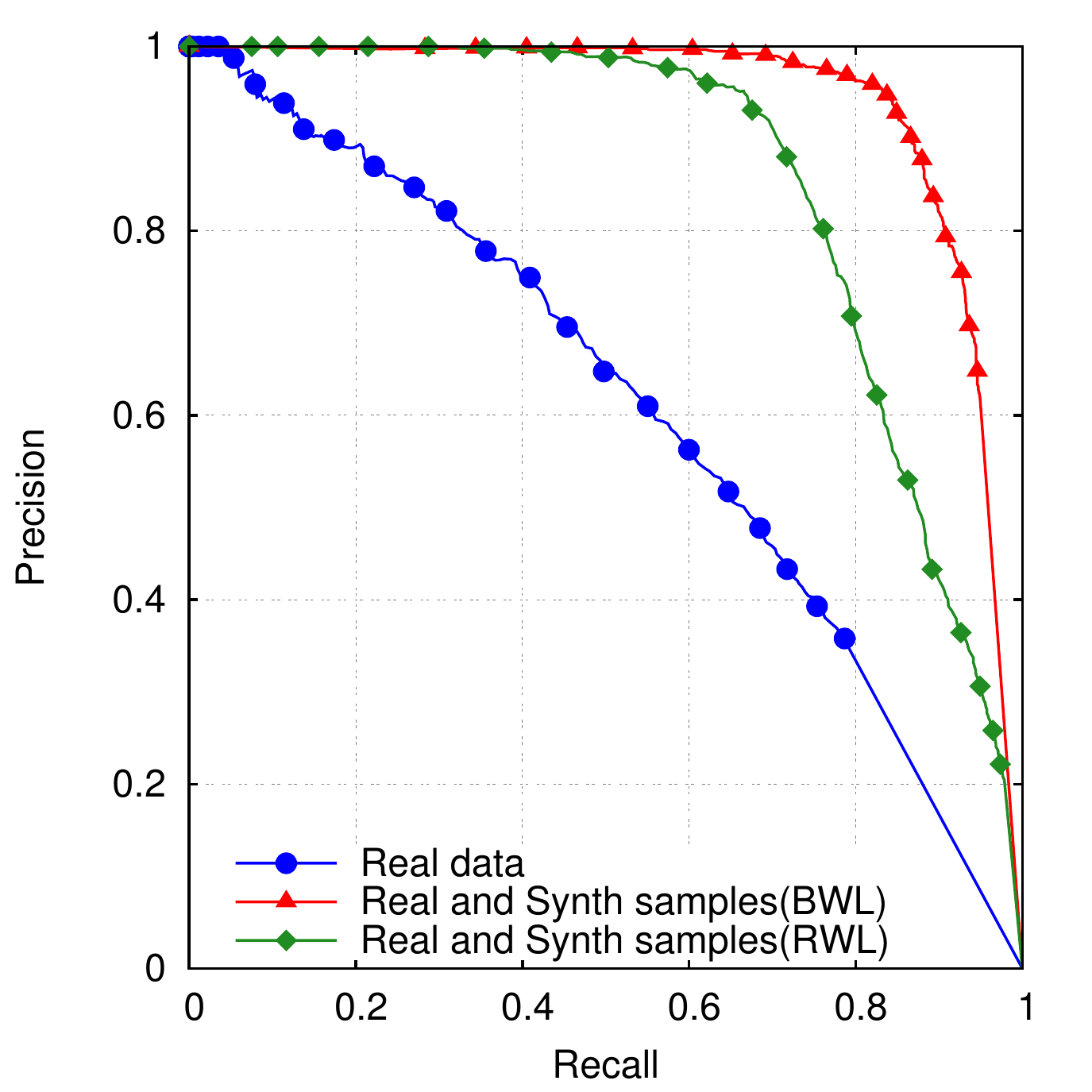} &
    \includegraphics[width=\threeim]{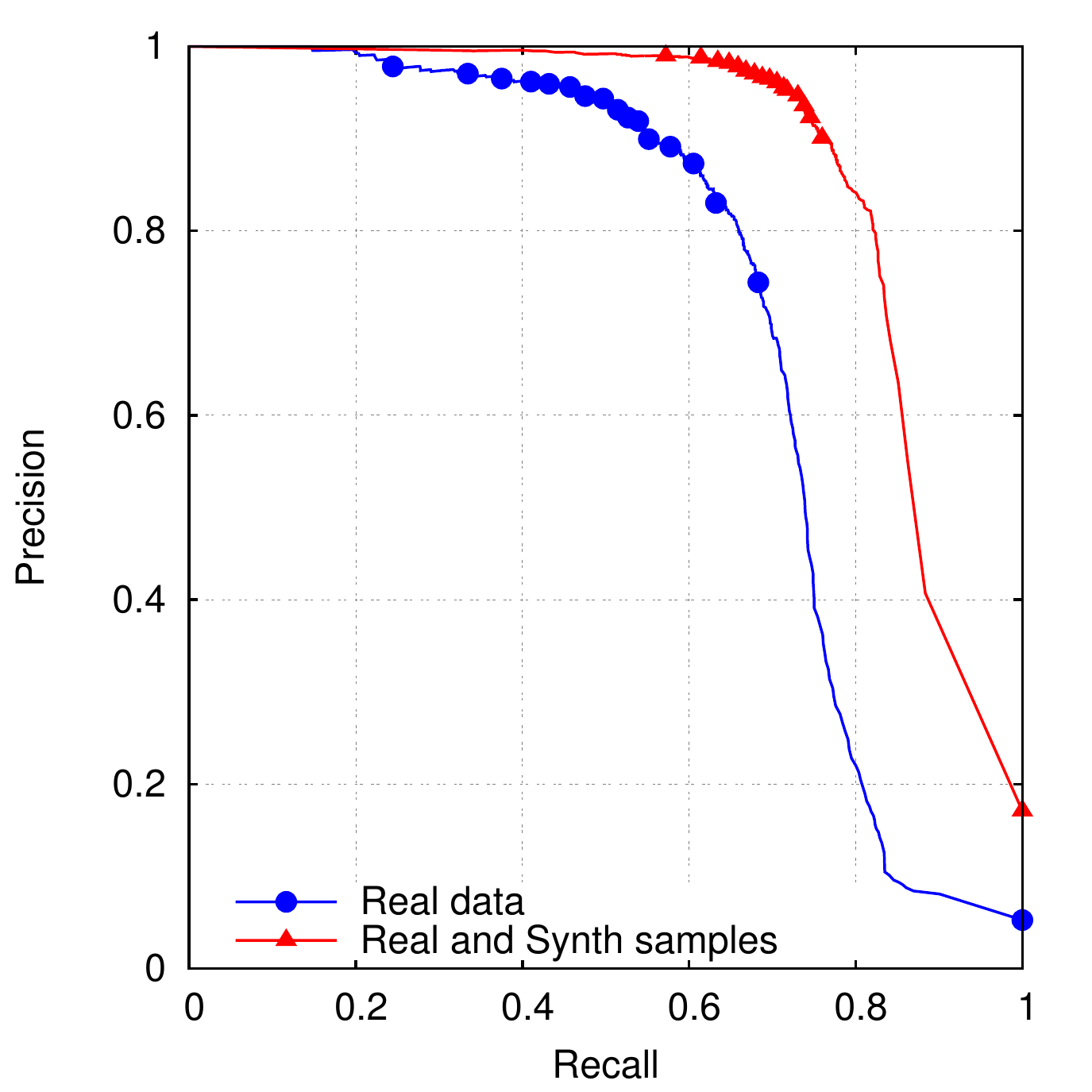} \\
    DPM and $d_\HOG(.,.)$ & AdaBoost and $d_\WL^{L,R}(.,.)$  & CNN and $d_\CNN(.,.)$ \\
  \end{tabular}
  \caption{\pascal{Comparing  the accuracy  of the  detectors trained  with real
      images only  (blue) against  those trained using  both synthetic  and real
      images (red or green).  AdaBoost  in particular does not perform very well
      with only the real images because  100 samples are not enough to train the
      algorithm  to  detect  3  different  kinds  of  aircrafts.   Nevertheless,
      introducing synthetic images lets us  enrich the training set to the point
      where performance  improves substantially in  all cases.  In  the AdaBoost
      case, the red  curve is obtained by using  the $d_\WL^{L}(.,.)$ similarity
      measure and the green one the $d_\WL^{R}(.,.)$ similarity measure.}}
  \label{fig:av_comp}
\end{figure*}

The test images come from 8 video sequences, one of which contains 5,000 frames,
while the others are made of 500 frames. These sequences show different types of
aircrafts  flying   in  different  environments  and   weather  conditions.   In
Table~\ref{tab:av_comp}  and Fig.~\ref{fig:av_comp},  we  compare results  using
real images only against an optimal combination of real and synthetic images.

Using the  detector trained on both  real and synthetic images  we achieve about
$90\%$ detection  accuracy, as opposed  to approximately $65\%$ when  using real
images only.  This  large improvement can be explained by the  fact that we have
only 100 real  images containing three different models,  while we generated 100
images  for each  real seed  image, which  results in  total in  10,000 positive
examples.   Table~\ref{tab:av_comp} illustrate  the  best accuracy  one can  get
varying the number of synthetic samples being added to the training set.  Sample
detections are shown in Fig.~\ref{fig:Detections_Plane}, which also depicts some failure cases.

\begin{table*}
  \centering
  \ra{1}
  \begin{tabular}{@{}lcccccc@{}}
    \toprule
    &&\multicolumn{5}{c}{\scriptsize{Similarity measure:}} \\
    \cmidrule{3-7}
    & \scriptsize{Using real} & \multirow{2}{*}{$d_\Eucl(.,.)$} &
    \multirow{2}{*}{$d_\HOG(.,.)$} & \multirow{2}{*}{$d_\WL^R(.,.)$} & \multirow{2}{*}{$d_\WL^L(.,.)$} & \multirow{2}{*}{$d_\CNN(.,.)$}  \\
    &\scriptsize{images only}&&&&&\\
    \midrule
    \scriptsize{Detection method:}&\multicolumn{6}{c}{\scriptsize{Average precision:}}\\
    DPM      & 0.79 & 0.83 & \bf{0.88} & 0.85 & 0.86 & 0.81 \\
    AdaBoost & 0.65 & 0.75 & 0.84 & 0.87 & \bf{0.92} \comment{0.94}& 0.73 \\
    CNN      & 0.72 & 0.75 & 0.85 & 0.70 & 0.83 &  \bf{0.88} \\
    \bottomrule
  \end{tabular} 
  \caption{\label{tab:av_comp} Comparing  average precisions for  each detection
    method when  the optimal  number of  synthetic images is  used for  the {\bf
      Aircraft  dataset}.  Each detection  method performs  best when  using the
    corresponding similarity function.}
\end{table*}

\begin{figure*}
  \centering
  \begin{tabular}{ccc}         
  \multicolumn{3}{c}{Detections} \\
    \includegraphics[width=\threeim]{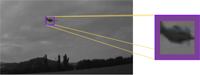} &
    \includegraphics[width=\threeim]{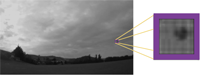}  &
    \includegraphics[width=\threeim]{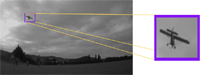}  \\
    \includegraphics[width=\threeim]{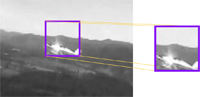}  &
    \includegraphics[width=\threeim]{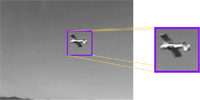}  &
    \includegraphics[width=\threeim]{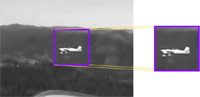}  \\
    \includegraphics[width=\threeim]{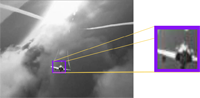}  &
    \includegraphics[width=\threeim]{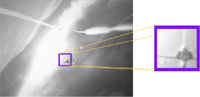}  &
    \includegraphics[width=\threeim]{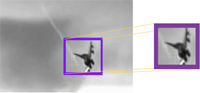}  \\
      \multicolumn{3}{c}{Missed and False Detections} \\ 
    \includegraphics[width=\threeim]{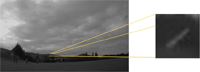} &
    \includegraphics[width=\threeim]{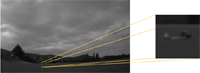}  &
    \includegraphics[width=\threeim]{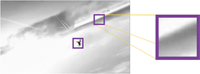}  \\
  \end{tabular}
  \caption{Examples of detections and some errors made by the detector, trained on both real and synthetic samples, and evaluated on the Aircraft dataset.}
  \label{fig:Detections_Plane}
\end{figure*}

\comment{
\newcommand{\dpsize}{1in}

\begin{figure*}
  \centering
  \begin{tabular}{cccc}         
    \includegraphics[width=\dpsize]{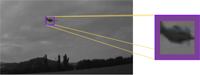} &
    \includegraphics[width=\dpsize]{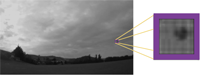}  &
    \includegraphics[width=\dpsize]{./_suppl_material/detections/av_1_1_98.png}  &
    \includegraphics[width=\dpsize]{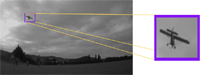}  \\
    \includegraphics[width=\dpsize]{./_suppl_material/detections/av_3_76_cr.png}  &
    \includegraphics[width=\dpsize]{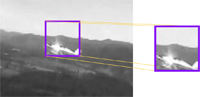}  &
    \includegraphics[width=\dpsize]{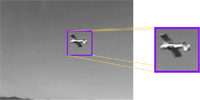}  &
    \includegraphics[width=\dpsize]{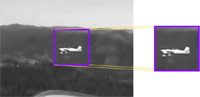}  \\
    \includegraphics[height = 0.75in,width=\dpsize]{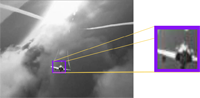}  &
    \includegraphics[height = 0.75in,width=\dpsize]{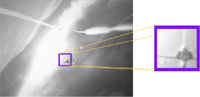}  &
    \includegraphics[height = 0.75in,width=\dpsize]{./_suppl_material/detections/av_6_5.png}  &
    \includegraphics[height = 0.75in,width=\dpsize]{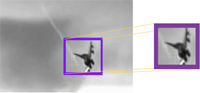}  \\    
  \end{tabular}
  \caption{Examples of detections made by the classifier trained on both real and synthetic samples on various aircraft and drone video sequences.}
  \label{fig:Detections_Plane}
\end{figure*}
}

\subsection{Comparing against another recent Image-Based Synthesis Approach}
\label{PVOC_comp}

\pascal{We compare here our approach with the one of~\cite{Rematas14}, which was
  applied to car detection on the PASCAL  VOC car dataset.  Like ours, it uses a
  CAD model of  the target object and seed real  images.  Its main contributions
  are the estimation of the material properties of every car component in a real
  image  and the  exploitation of  this  information to  generate new  synthetic
  views, which  are then used to supplement  real ones to train  a DPM detector.
  This requires registration  of the model so that it precisely  fits the car in
  the image and the image texture can then back-projected onto the car model, so
  that material properties can be assigned to each visible part of the model. To
  generate a new view, the model is  rotated in 3D and re-projected in the scene,
  which includes the ground plane and  the background plane.  This ends up making
  some previously  invisible parts of  the car visible. Material  properties for
  these  newly visible  parts are estimated  using a  weighted sum  of the
  properties  of  the  parts   whose  material  properties  have  already  been
  estimated.}

\begin{figure*}
  \centering
  \begin{tabular}{cccc}
    \includegraphics[height=0.85in]{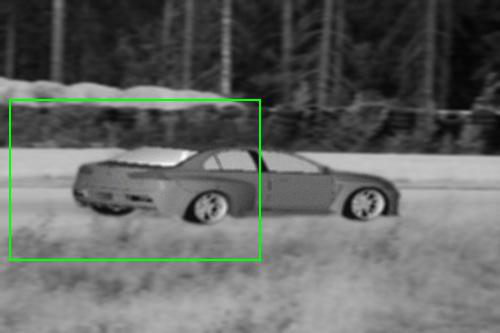}  &
    \phantom{abc}                                                  &
    \includegraphics[height=0.85in]{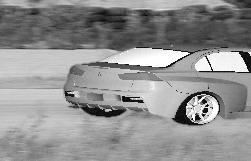}  &
    \includegraphics[height=0.85in]{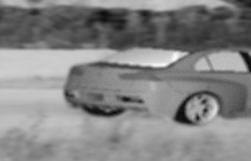} 
    \\ \includegraphics[height=0.85in]{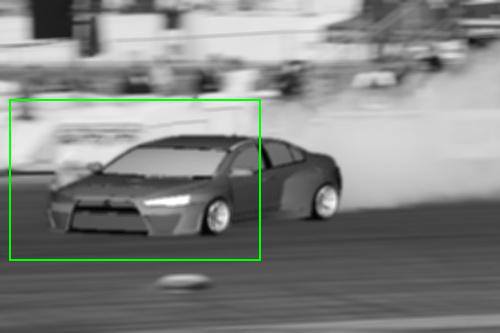} 
    &                         \phantom{abc}                        &
    \includegraphics[height=0.85in]{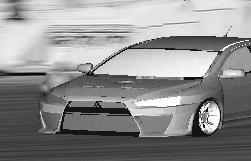}    &
    \includegraphics[height=0.85in]{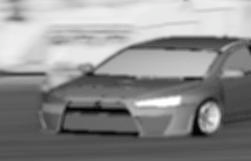}   \\ (a)
    & & (b) & (c)
  \end{tabular}
  \caption{\label{fig:car_gray} (a)  Sample synthetic images of  cars generated by
    our  approach.  (b, c) Patches  extracted  from  these images  emphasizing  the
    importance of the boundary and motion blurring effects.  }
\end{figure*}

Since, our  algorithm requires  background images in  addition to the  model and
seed real images,  we derived them from  the seed images by cutting  out the car
and filling the empty  space using content aware texture filling~\cite{Ruzic12}.
Sample images are shown in Fig.~\ref{fig:car_gray}.

\begin{figure*}
    \centering
    \begin{tabular}{ccc}
      \includegraphics[height=0.95in]{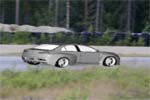} &
      \includegraphics[height=0.95in]{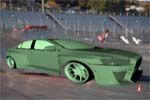}  &
      \includegraphics[height=0.95in]{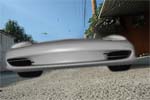}  \\
      \includegraphics[height=0.95in]{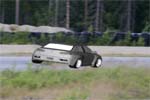}  &
      \includegraphics[height=0.95in]{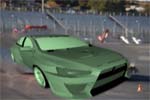}  &
      \includegraphics[height=0.95in]{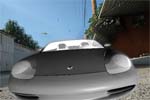}  \\
      (a) & (b) & (c) \\
    \end{tabular}
    \caption{Sample  color synthetic images  of cars,  generated by  our system.
      (b) does not look very  realistic, because the same properties are applied
      to  all the  components of  the car.   \pascal{A more  sophisticated model
        could be used  to address this issue.  However realism  does
        not seem to be
        critical for  our purposes since  our simplistic model is  sufficient to
        outperform~\cite{Rematas14}}.}
    \label{fig:car_synth}
  \end{figure*}

\begin{figure*}
    \centering
    \begin{tabular}{ccc}
      \multicolumn{3}{c}{Correct Detections} \\ 
      \includegraphics[width=\threeim]{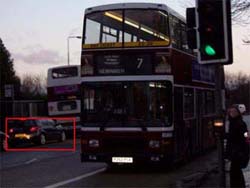} &
      \includegraphics[width=\threeim]{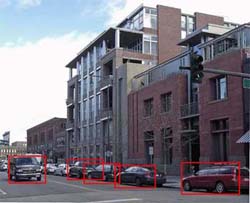}  &
      \includegraphics[width=\threeim]{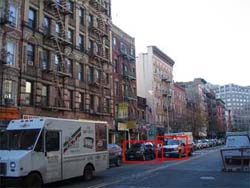}  \\
      \includegraphics[width=\threeim]{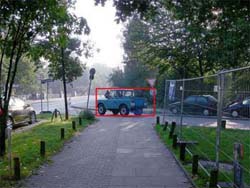}  &
      \includegraphics[width=\threeim]{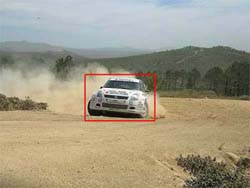}  &
      \includegraphics[width=\threeim]{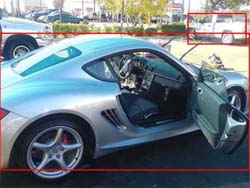}  \\
      \multicolumn{3}{c}{Mis-detections} \\ 
      \includegraphics[width=\threeim]{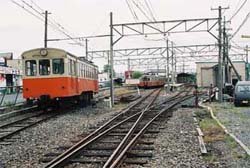} &
      \includegraphics[width=\threeim]{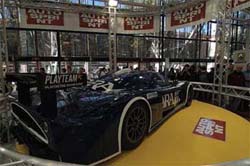}  &
      \includegraphics[width=\threeim]{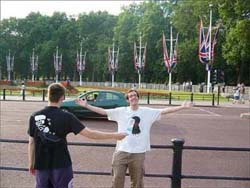}  \\
    \end{tabular}
    \caption{Sample detections made by the 5 component DPM, trained on the full
      real VOC dataset, supplemented by synthetic data, generated by our
      method. Last row shows mis-detections. (best seen in color) }
    \label{fig:car_det}
  \end{figure*}

The  images of the  Pascal VOC  dataset being  in color,  for a  fair comparison
against~\cite{Rematas14} that  exploits this fact,  we extended our  approach to
color images  by simply optimizing on  the $\Theta$ parameters on  the three RGB
channels independently, which  yields three sets of parameters  for every image.
These parameters are  then used to generate separate  synthetic images for every
channel, and  finally combined in one  RGB image.  We  vary the pose of  the car
model,  but  also the  direction  of  the light  source,  which  cannot be  done
with~\cite{Rematas14}.   The  results  of  this  combination  are  presented  in
Fig.~\ref{fig:car_synth}.  The car  in Fig.~\ref{fig:car_synth}(b) does not look
very realistic, because the same properties are applied to all the components of
the car.  A more sophisticated model  would solve this issue, however we already
obtain satisfying results using this  simplistic rendering, which confirms that producing visually pleasing synthetic images is not a primary requirement.  Some detections made by the  5 component  DPM framework,  trained on  both real  and synthetic  data are presented in the Fig.~\ref{fig:car_det}. 

\pascal{Table~\ref{tab:pascal_comparison}          shows         that         we
  outperform~\cite{Rematas14} even  though our approach  was originally designed
  to  generate  small  image  patches  centered  on  the  target  object.
  Furthermore,  as  shown  in  the   previous  sections,  it  is  applicable  to
  low-resolution  images with  very limited  texture, for  which the  the method
  of~\cite{Rematas14} is not  well adapted. Furthermore, if we  don't use color,
  the performance drops by only 1-2\%, which is not very large.  }

\begin{table*}
  \centering
  \ra{1}
  \begin{tabular}{@{}clccc@{}}
    \toprule
    Test set & Training set \phantom{abcaaaabcabcabcabcabcabcabcabc} &\multicolumn{3}{c}{Avg. Precision} \\
    \midrule
    \multicolumn{2}{r}{ \scriptsize{Number of components:}} & $N$ = 3 & $N$ = 4 & $N$ = 5 \\
    \midrule
    \multirow{6}{*}{\rotatebox{-90}{VOC 2007}} & {Side} & 16.2 & 18.4 & 16.7 \\
    & {Side+Synth~\cite{Rematas14}} & 30.2 & 31.4 & 33.2 \\
    & {Side+Synth (Our method)} & {\bf 35.1} & {\bf 37.9} & {\bf 38.0} \\
    \cmidrule{2-5}
    & {Full} & 51.7 & 53.4 & 50.7 \\
    & {Full+Synth~\cite{Rematas14}} & 50.2 & {\bf 53.1} & 50.9 \\
    & {Full+Synth (Our method)} & {\bf 52.1} & 52.9 & {\bf 55.3} \\
    \bottomrule
  \end{tabular} 
  \caption{\label{tab:pascal_comparison} Comparing  the performances of  the DPM
    detector  on  the  PASCAL VOC  car  dataset  when  trained with  the  method
    of~\cite{Rematas14} and our  method as a function of $N$,  the number of DPM
    components. The  performance of  the detectors trained  using only  the real
    images is also given for reference. }
\end{table*}

\section{Conclusion}

We have shown that by properly optimizing  the parameters of a very simple rendering
pipeline,  we  can generate  synthetic  images  that significantly  improve  the
performance of  an object detector when  used for  training.  We  believe our
parameter optimization scheme is a powerful  tool to manage the large numbers of
parameters a more complex rendering pipeline  could have. It therefore opens new
doors towards  the use  of sophisticated  Computer Graphics  and post-processing
effects to generate images even closer to real ones, and to relax the cumbersome
need for large numbers of real images to train Computer Vision methods.

\bibliography{string,vision}

\newpage

\parpic{\includegraphics[width=1in,clip,keepaspectratio]{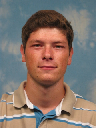}}
\noindent {\bf Artem Rozantsev} joined EPFL in 2012 as Ph.D. candidate at CVLab, under the supervision of Prof. Pascal Fua and Prof. Vincent Lepetit. He received his Specialist degree in Mathematics and Computer Science from Lomonosov Moscow State University. His main research interests include object detection, synthetic data generation and machine learning.

\vfill

\parpic{\includegraphics[width=1in,clip,keepaspectratio]{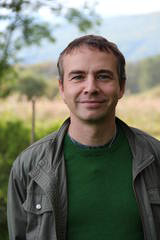}}
\noindent{\bf Vincent Lepetit}
is a Professor at the Institute for Computer Graphics and Vision, TU Graz and a Visiting Professor at the Computer Vision Laboratory, EPFL. He received the engineering and master degrees in Computer Science from the ESIAL in 1996. He received the PhD degree in Computer Vision in 2001 from the University of Nancy, France, after working in the ISA INRIA team. He then joined the Virtual Reality Lab at EPFL as a post-doctoral fellow and became a founding member of the Computer Vision Laboratory. His research interests include vision-based Augmented Reality, 3D camera tracking, Machine Learning, object recognition, and 3D reconstruction.

\vfill

\parpic{\includegraphics[width=1in,clip,keepaspectratio]{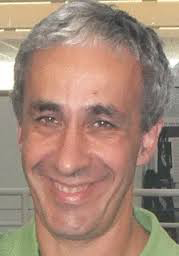}}
\noindent{\bf Pascal Fua} received an  engineering degree from Ecole Polytechnique,  Paris, in 1984 and  the Ph.D.  degree  in Computer  Science  from the  University  of Orsay  in 1989.  He joined EPFL  in 1996  where he  is now  a Professor  in the  School of Computer and Communication Science. Before  that, he worked at SRI International and at  INRIA Sophia-Antipolis as  a Computer Scientist. His  research interests include shape modeling  and motion recovery from images,  analysis of microscopy images,  and Augmented  Reality. He  has (co)authored  over 150  publications in refereed journals  and conferences.  He is an  IEEE fellow  and has been  a PAMI associate editor.  He often serves as  program committee member,  area chair, or program chair of major vision conferences.
\vfill

\end{document}